\documentclass[12pt,a4paper]{article}
\usepackage{times,slashbox}
\usepackage{epsfig}
\usepackage{graphicx,xcolor}
\usepackage{amsmath,amssymb,amsthm,bbm}

\usepackage{enumitem}
\usepackage{booktabs}  
\usepackage{algorithm}
\usepackage{algpseudocode}

\def\P{{\mathbb P}}
\def\E{{\mathbb E}}
\def\R{{\mathbb R}}

\def\ncal{{\mathcal N}}
\def\sphere{\mathbb{S}^{N-1}}

\newcommand{\normal}[2]{\mathcal{N}({#1},{#2})}
\newcommand{\gxy}{g_{\delta}}
\newcommand{\gxn}{g_\ncal}

\def\Hcal{{\mathcal H}}

\definecolor{darkblue}{rgb}{0.1,.1,0.8}

\newcommand{\preal}{\P^{\text{real}}}
\newcommand{\indep}{\perp\!\!\!\perp}

\newcommand{\platent}{{\mathbb L}_z}
\newcommand{\probalaws}{\mathcal{P}(\Omega)}
\def\beq{\begin{equation}}
\def\eeq{\end{equation}}
\def\beqn{\begin{eqnarray}}
\def\eeqn{\end{eqnarray}}

\newtheorem{theorem}{Theorem}
\newtheorem{proposition}[theorem]{Proposition}
\newtheorem{remark}[theorem]{Remark}
\newtheorem{corrolary}[theorem]{Corrolary}
\newtheorem{lemma}[theorem]{Lemma}

\usepackage{lineno,hyperref}

\title{Radon Sobolev Variational Auto-Encoders}

\author{Gabriel Turinici\footnote{\text{https://turinici.com, Gabriel.Turinici@dauphine.fr}}
		 \\
Université Paris Dauphine - PSL Research University\\
CEREMADE, \\ Place du Marechal de Lattre de Tassigny, Paris 75016, FRANCE}

\date{April 02, 2021}

\begin{document}
\maketitle
\begin{abstract}
The quality of  generative models (such as Generative adversarial networks and Variational Auto-Encoders) 
depends heavily on the choice of a good 
probability distance. However some popular metrics like the Wasserstein or the  Sliced Wasserstein distances, the Jensen–Shannon divergence,  the Kullback–Leibler divergence, 
lack convenient properties such as (geodesic) convexity, fast evaluation and so on. 
To address these shortcomings, we introduce a class of distances that have built-in convexity. 
We investigate the relationship with some known paradigms (sliced distances -a synonym for Radon distances -, 
reproducing kernel Hilbert spaces, energy distances).
The distances are shown to possess fast implementations and
are included  in an adapted Variational Auto-Encoder 
termed Radon Sobolev Variational Auto-Encoder (RS-VAE) which produces high quality results
on standard generative datasets.

Keywords: Variational Auto-Encoder; Generative model; Sobolev spaces; 
	Radon Sobolev Variational Auto-Encoder;
\end{abstract}

\section{Introduction}

Deep neural networks used as generative models are of high interest in a large array of application domains
~\cite{kingma2013autoencoding,originalGAN14,tolstikhin2017wasserstein,arjovsky2017wgan}. 
However they come at the price of a more intricate architecture and convergence patterns than supervised networks.

The goal of generative models is to design a procedure to 
sample from a target probability law
$\preal$ using a dataset of available samples $Y_1$, ..., $Y_L \sim \preal$ (the number 
$L$ of samples is large but fixed)\footnote{Throughout the paper $\sim$ denotes equality in law.}. 

One of the most used and efficient architectures are the Generative Adversarial Networks (GANs)\cite{originalGAN14,arjovsky2017wgan,Deshpande_2019_CVPRmax} ; GANs come in the form of a dual
net: a generator and a discriminator, whose joint convergence was shown to pose problems, addressed in
late variants (see WGAN, SWGAN, etc.). As generators deal with probability laws, 
the choice of the distance used to quantify the closeness of a candidate turns out to be  
of critical importance. This is even more visible for Variational Auto-Encoders (VAE) (see \cite{kingma_introduction_2019} for a recent introduction
	to the subject)
that use the distance as a part of the loss functional (i.e., not in the dual form as most of the GANs do, see \cite{arjovsky2017wgan} for the relationship 
	between  the 
	Discriminator part of a GAN  and the dual formulation of the Wasserstein distance).

A VAE has two stages: an encoder stage $E_{\theta_e}(\cdot)$ indexed
by the parameters $\theta_e$ of the encoding network, and a decoding network
$D_{\theta_d}$. The networks are fitted in order to satisfy two goals: first the reconstruction
error $D_{\theta_d}(E_{\theta_e}(Y_k)) -Y_k$ is to be minimized over all available samples $Y_k$; this 
is usually implemented by minimizing the mean square error.
On the other hand, the second requirement is to minimize the mismatch
between the encoded empirical distribution  $(E_{\theta_e})_* \preal$ and a 
 target law $\platent$ (here the symbol "$*$" means the image of the distribution 
 on the right with respect to the 
 mapping on the left, also called the push-forward map).
 
  In the generation phase 
 one takes as input 
 samples $z$ from the law $\platent$ and 
 maps them through the decoding function $D_{\theta_d}$ in order to generate new samples not seen in the dataset $Y_1, ..., Y_L$.

Let $\Omega$ be the set of all possible values of the latent sample $z$ and $\probalaws$ 
the set of all probability laws on $\Omega$; then, crucial to the VAE is 
the distance $d(\cdot,\cdot)$ acting on
 (possibly a subset of) $\probalaws^2$ that measures the mismatch between 
 $(E_{\theta_e})_* \preal$ and the latent distribution $\platent$. 
 The computation of the distance $d$ is usually difficult
and in many cases one resorts to approximations.

The goal of this paper is to present a class of distances relevant in practice, 
easy to compute and that have built-in convexity (thus easing the minimization procedure).

Our proposal builds on several ideas: on one hand the "kernel trick" that enables the computation of 
functional quantities on Euclidian spaces; on the other hand the "sliced distances"  
that average, in a sense to be made precise, the distances of projections
to one-dimensional subspaces. We use the "Radon" term instead of "sliced" 
 to remind that the basic ingredient is the projection on one-dimensional spaces
and to draw connections with earlier literature on Radon transform,  
see \cite{radon_uber_1917,xraytransform38,natterer_mathematics_1986}, \cite[Chapter 2]{book_Filippo}.
 The third and last ingredient is a Hilbert space of regular functions. The regularity has often
 been invoked in relation with GANs for instance in the Wasserstein GAN form (that use Lipschitz functions). We encode the smoothness of a function through a parameter $s\ge 0$ of 
 some Sobolev space $H^s$ or $\dot{H}^s$
(see definition in sections~\ref{sec:sobolevhs} - \ref{sec:sobolevhdot}), larger $s$ meaning smoother.
 
\section{Desirable properties of a distance}

We discuss in this section some specific properties that a distance should have in order to be suitable 
for use in generative models.

\subsection{Convexity}

A property easy to understand is the convexity; it is not a hard requirement but it certainly helps the convergence 
and robustness with respect to perturbations.

\begin{figure}
	   \includegraphics[height=5cm]{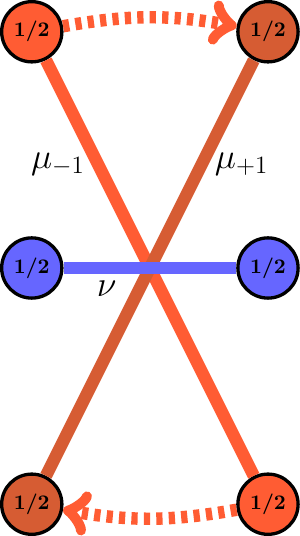} \hfill 
		\includegraphics[width=0.5\linewidth]{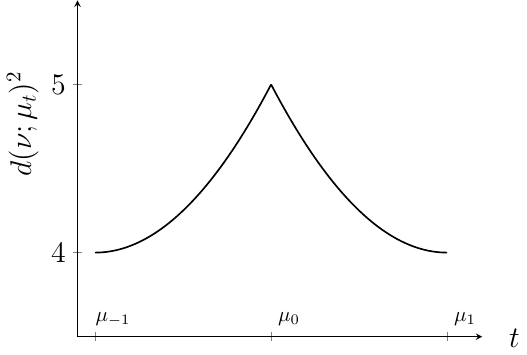}
	\caption{
		{\bf Left:}
		 The geodesic line 	$(\mu_t)_{t \in [-1,1]}$  between $\mu_{-1}$ and $\mu_{-1}$ for the $W_1$ distance consists in
			moving the top-left mass to the right and simultaneously the bottom-right mass to the left, at constant speed following the
			dotted trajectory.
		{\bf Right:} We see that in this case, the square of the distances to some reference measure $\nu$ (in blue) evolves in a strongly nonconvex way.
	}  \label{fig:nonconvexWass}
	\end{figure}

Let us be more clear what convexity means in a space of probability laws. Suppose a metric $d$ is given on $\probalaws$; then
given two distributions $\mu_{-1}$ and $\mu_1$ one can consider the geodesic $\mu_t$ starting at $\mu_{-1}$ and ending in $\mu_1$. In 
 Euclidian spaces this would be just a straight line\footnote{Here and in the sequel, 
 	we use the terminology of "straight line" to designate a segment relative to the affine structure i.e., a linear combination of its endpoints
 	which is well defined in any vector space.} but in general this is not the case~\cite{book_Filippo}.
Take for instance the Wasserstein metric $d_{W}$ used for instance in WGANs (see~\cite{arjovsky2017wgan} for details concerning 
the definition of such metrics), and the following example (adapted from~\cite[page 275]{book_Filippo}):
consider the geodesic 
$\mu_t=\frac{1}{2} \delta_{(t,2)} + \frac{1}{2} \delta_{(-t,-2)}$
linking $\mu_{-1}$ and $\mu_1$  and the "target" law $\nu=\frac{1}{2} \delta_{(1,0)} + \frac{1}{2} \delta_{(-1,0)}$. Then the 
	distance squared from $\nu$ to the points $\mu_t$ on the geodesic is given by: 
	$ d(\nu,\mu_t)^2 = 4+ \min((1-t)^2,(1+t)^2)$, which is not convex and in particular has two local minima, see 
	Figure~\ref{fig:nonconvexWass}. But, the distance squared function is an important ingredient 
	of the many loss functions and the additional non-convexity added because of using the Wasserstein distance  
	will make  the optimization more difficult.

\subsection{Nonlocal computability} \label{sec:nonlocalcalculability}

Another suitable property of a distance is related to its computability. It is very convenient to deal with distances that can be computed 
 with as little information as possible concerning the "local" (thus specific and changing) properties of the distributions 
 in the argument. 
For instance, in a metric space, there may be possible to express the distance between an external point and an arbitrary point on a geodesic segment using only the distance 
between the external point and the two extremal points of the geodesic segment; such a 
formula is very practical (e.g., when computing the distance is costly)
and has been used in many works~\cite{giglibook}
 \footnote{Here the gain in efficacy is given by the fact that three quantities are enough to compute a infinity of distances. As we will see later, the distance 
 from a Dirac mass to another Dirac mass will be translation invariant, another convenient property.}. We will consider in particular the following property:
\beqn \label{eq:ddd_id}
& \ & 
\!\!\!\!\!\!\!\!
\text{For any geodesic } \mu_t:[0,1] \to \probalaws \text{ and any } \nu \in \probalaws :
\nonumber \\ & \ & 
\!\!\!\!\!\!\!\!
\!\!\!\!\!\!\!\!
d^2(\nu,\mu_t) = (1-t) \cdot d^2(\nu,\mu_0) + t \cdot d^2(\nu,\mu_1)
 - t(1-t)\cdot d^2(\mu_0,\mu_1), \forall t\in [0,1]. \label{eq:parallelogram}
\eeqn
 Although at first cryptic, in a Hilbert space this is nothing more than, for instance, the parallelogram identity 
 $\| x+y \|^2 = 2 \cdot \|x\|^2 + 2 \cdot \|y\|^2 - \|x-y\|^2$ expressed at $\nu=0$, $\mu_0=x$, $\mu_1=y$, $t=1/2$.
In fact~\eqref{eq:ddd_id} is satisfied in any Hilbert space because the geodesics are straight lines. 
For instance the space $L^2$ of square integrable functions is a good example, while the space $L^4$ 
of functions with finite norm  $\|f \|_{L^4} = \left(\int f^4 \right)^{1/4}$ does not fulfill the condition.
 Reciprocally, if a metric space is 
endowed with an algebraic operation compatible with the distance and the above identity, it can then be isometrically embedded into a  Hilbert space.\footnote{
	This conclusion is relatively easy to obtain when the space has a Banach (i.e. complete, normed) structure and $t \in [0,1]$ in \eqref{eq:parallelogram} is arbitrary. 
	When  \eqref{eq:parallelogram} only works with $t=1/2$ this is the Jordan-von Neuman embedding theorem \cite{jordan_neuman_inner_product}. 
	When the overall space is a more general metric space,
	the proof is more involved and dates back to 
	Blumenthal \cite{blumenthal_theory_1953} while the form most close to our setting is that of M.M. Day, see 
	\cite{day_criteria_1959} Corollary 2 page 100, relying on the 
	so called "queasy euclidean fourpoint property" hypothesis	denoted by "(e4pp-3)" (page 97, same reference) which is a rewriting of  \eqref{eq:parallelogram}.
}

In practice this allows to have the following:

\begin{proposition}  \label{prop:computation}
	Let $\Omega$ a subdomain of $\R^N$ and 	
	$(X,d)$ a metric space containing all Dirac masses $\delta_x$, for all $x\in \Omega$. Then if $(X,d)$ satisfies
property~\eqref{eq:ddd_id} on any straight line $\mu_t = (1-t)\mu_0 + t \mu_1$ then
for any $x_1,..., x_K$, $y_1,..., y_J \in \Omega$~:
\begin{multline}
d \left( \frac{1}{K} \sum_{k=1}^K \delta_{x_k}, \frac{1}{J} \sum_{j=1}^J \delta_{y_j} \right)^2=  
\frac{1}{K\cdot J} \sum_{k=1}^K \sum_{j=1}^J \!\!  d(\delta_{x_k},\delta_{y_j})^2 
- \frac{1}{2 K^2} \sum_{k=1}^K\sum_{k'=1}^K \!\!  d(\delta_{x_k},\delta_{x_{k'}})^2 
\\
-\frac{1}{2 J^2} \sum_{j=1}^J\sum_{j'=1}^J \!\!  d(\delta_{y_j},\delta_{y_{j'}})^2.
\label{eq:distsumdiracs}
\end{multline}
\end{proposition}
Here and in the following, the algebraic expression $\alpha \mu_0+ \beta \mu_1$ for $\alpha, \beta \in \R $ and $\mu_0, \mu_1 \in X$ 
	is defined as the mixture of the measures $\mu_0$ and $\mu_1$ 
	 which, when
	 integrated against some arbitrary function $f:\Omega \to \R$ gives as result
	 $ \int_{\Omega} f(\cdot) (\alpha \mu_0 + \beta \mu_1) = \alpha \int_{\Omega} f(\cdot) \mu_0 + \beta \int_{\Omega} f(\cdot) \mu_1$. Same extends for finite sums. In particular when
	 $\mu_0$ and $\mu_1$ are Dirac masses 	 $\mu_0 = \delta_y$ and $\mu_1=\delta_y$ :	 
	 $ \int_{\Omega} f(\cdot) (\alpha \delta_x + \beta \delta_y) = \alpha f(x) + \beta f(y) $.

\begin{proof}
	We proceed by recurrence over $K+J$ using equation~\eqref{eq:ddd_id}. Since $K,J \ge 1$ the first possible value is $2$ when $K=J=1$ and the proof is an identity 
	because the last two sums in \eqref{eq:distsumdiracs} consist each of a single, null, term. When $K+J>2$ one of $K$ or $J$ is greater than $1$, suppose $J\ge 2$.
We use hypothesis~\eqref{eq:ddd_id} for $\nu = \frac{1}{K} \sum_{k=1}^K \delta_{x_k}$, $\mu_0=\delta_{y_1}$, $\mu_1= \frac{1}{J-1} \sum_{j=2}^J \delta_{y_j}$, $t=1-1/J$:
\begin{multline}
d \left( \frac{1}{K} \sum_{k=1}^K \delta_{x_k}, \frac{1}{J} \sum_{j=1}^J \delta_{y_j} \right)^2=  
\frac{1}{J}  d \left( \frac{1}{K} \sum_{k=1}^K \delta_{x_k}, \delta_{y_1} \right)^2
\\
+
\frac{J-1}{J}  d \left( \frac{1}{K} \sum_{k=1}^K \delta_{x_k}, \frac{1}{J-1} \sum_{j=2}^J \delta_{y_j} \right)^2
- \frac{J-1}{J^2}  d \left( \delta_{y_1}, \frac{1}{J-1} \sum_{j=2}^J \delta_{y_j} \right)^2 .
\label{eq:distsumdiracsP} 
\end{multline}
All three terms can now be computed through the recurrence hypothesis because the number of Dirac masses involved are 
$K+1$, $K+J-1$ and $J$ respectively, all strictly below $K+J$. 
By recurrence, the result will be a sum containing the terms 
$d \left( \delta_{x_k}, \delta_{x_{k'}} \right)^2$, 
$d \left( \delta_{y_j}, \delta_{y_{j'}} \right)^2$ and
$d \left( \delta_{x_k}, \delta_{y_j} \right)^2$ 
for $1 \le k, k' \le K$, $1\le j, j'\le J$. 
For cross-terms $d \left( \delta_{x_k}, \delta_{y_j} \right)^2$
we only need to investigate one of such terms, for instance $d \left( \delta_{x_1}, \delta_{y_1} \right)^2$ because all other such terms will be similar by symmetry: the
distance $d \left( \delta_{x_1}, \delta_{y_1} \right)^2$
 only appears in the first quantity
in the first line of the above formula. By recurrence its coefficient will be $  \frac{1}{J} \cdot \frac{1}{K \cdot 1} = \frac{1}{K J}$. Finally, 
for terms like $ d(\delta_{y_i},\delta_{y_j})^2$ we take as example 
 $d \left( \delta_{y_1}, \delta_{y_2} \right)^2$, only produced by the last term in equation  \eqref{eq:distsumdiracsP}; its  coefficient is $- \frac{J-1}{J^2} \cdot  \frac{-2}{1 \cdot (J-1)} = - \frac{2}{J^2}$ as expected.
\end{proof}
\begin{remark} A very similar proof shows that in general, for any $\alpha_1, ... \alpha_K, \beta_1, ..., \beta_J \ge 0$ with $\sum_{k=1}^K \alpha_k=1=\sum_{j=1}^J \beta_j$:
\begin{multline}
d \left(\sum_{k=1}^K \alpha_k \delta_{x_k}, \sum_{j=1}^J \beta_j \delta_{y_j} \right)^2=  
\sum_{k=1}^K \sum_{j=1}^J \alpha_k \beta_j d(\delta_{x_k},\delta_{y_j})^2 
\\
- \frac{1}{2} \sum_{k=1}^K\sum_{k'=1}^K \!\! \alpha_k \alpha_{k'}  d(\delta_{x_k},\delta_{x_{k'}})^2 
- \frac{1}{2} \sum_{j=1}^J\sum_{j'=1}^J  \beta_j \beta_{j'} d(\delta_{y_j},\delta_{y_{j'}})^2.
\label{eq:distsumdiracsab}
\end{multline}
\end{remark}

Proposition~\ref{prop:computation} is a  powerful tool because of two reasons: first it allows to compute the 
distance in an easy way, second this computation is in the form of expectations (replacing the sums by expectation over a discrete uniform law; 
see equation~\eqref{eq:energynorm} below for an example).

\section{Relationship with the literature} \label{sec:literature}

This work owes much to previous advances from the literature. 

Of course, first of all is the idea that a (W)GAN can be seen as minimizing some probability metric~\cite{arjovsky2017wgan}. They use
a Wasserstein metric which is similar to ours (see equation~\eqref{eq:relationwassh1} below and section~\ref{sec:comparison}) but 
 whose computation requires iterations and the enforcing of some special (Lipschitz type) constraints.

Starting from this difficulty, several solutions were proposed~: first the 
"sliced" distances, most used being the "sliced Wasserstein distance" appearing  in~\cite{Kolouri_2018_CVPR,Deshpande_2018_CVPR,kolouri2018sliced,Wu_2019_CVPRgenerative,Deshpande_2019_CVPRmax} 
to cite but a few. Our "Radon" terminology is but a synonym for "slicing" but will be used from now for coherence with previous literature see  \cite{radon_uber_1917,xraytransform38,natterer_mathematics_1986}.
Other forms of sliced distances have also been proposed, in particular in~\cite{CWAE} authors implement a VAE using a kernelized distance; note however that in
 full rigor the metric they used is not a distance because the smoothing parameter depends on the number of Dirac masses 
 present in the distributions. However it is a good metric that allows one to compute analytically the distance from a Dirac mass (and hence any finite support distribution)
 to a standard multivariate normal (distance used to converge to the latent distribution).

On the other hand there is the remarkable work on the energy distance by Szekely and al. (see~\cite{SZEKELY13} and related references)
that establishes an encouraging framework; contrary to the first group cited, here the approach is global; note 
that in~\cite{sobolevgan} the authors use Sobolev spaces but instead of using Radon (or "sliced") versions they use directly
 the overall space $H^s(\R^N)$; the inconvenience of  this space is that it is included in the space of continuous functions only 
 when $s \ge N/2$ which means that, since $N$ is large, a high regularity is required in order for the dual to contain Dirac masses. A close work is~\cite{cramergan} 
which implements a particular form of Radon Sobolev distance (for $H=\dot{H}$) in the framework of a GAN; 
as such it requires the use of a space of features that is to be optimized.

Finally, this work can be put in the more general framework of 
Hilbert space embeddings (see for instance~\cite[Theorem 21 p. 1551]{hilbertspaceembed10} 
and~\cite{AnnalsStat_endist13}) that give theoretical insights into the use of energy-type distances and relationship with  maximum mean discrepancies
(MMD) metrics.

\subsection{Contributions of this work}

This work proposes some novelties with respect to the literature that we detail below:

- first from a theoretical point of view, we introduce a novel procedure to construct a class of probability distances
taking into account the regularity of the test functions; note that this regularity is precisely what 
delimits for instance the total variation distance  (continuous functions) from the Wasserstein distance
(Lipschitz functions); 

- in particular the "energy distance" of Szekely et al. (see \cite{SZEKELY13}) is a particular member of this class; this distance has been
tested extensively outside the deep learning community;

- the distances are interesting computationally because one has quasi-analytic formulas for computing the distances
among sums of Dirac masses and from a sum of Dirac mass to the standard normal multivariate distribution; furthermore
some aysmptotic expansions (giving good results in practice) are proposed;

- we propose an adapted Variational Auto-Encoder (termed RS-VAE to recall that the distance used is in the proposed class)
 that obtains good results on  standard datasets.
 In particular the distance is seen to be a reliable proxy for other
 metrics (including Sliced Wasserstein and Cramer-Wold). Note that in practice it is difficult to compare with non-sliced VAEs because the distances, exact to the extent needed 
 to make meaningful comparisons, are difficult to obtain in high
 dimensional latent spaces. This comparison is for instance the approach used in other, not directly related, endeavors, see~\cite{karras2018progressive}.

\section{Theoretical results}

 We introduce below some basic notions concerning the Sobolev spaces $H^s(\R)$ and $\dot{H}^s(\R)$.

\subsection{Spaces $H^s$} \label{sec:sobolevhs}

Recall that $L^2(\Omega)$ (also denoted $L^2$ when there is no ambiguity) is the space of real functions $f$ defined on $\Omega$ such that 
$f^2$ is integrable. If $f \in L^2$ and its first derivative $\nabla f$ is also in $L^2$ then we say that $f$ belongs to the 
Sobolev space $H^1$; this construction can be iterated: if  
$f$ is  $m$-times differentiable and all (partial) derivatives of rank $m$ are in $L^2$ then  $f \in H^m$. 
Let us take $\Omega= \R^N$ and recall that the Fourier transform maps the derivation operator into
the multiplication by the (dual) argument\footnote{Such a property is especially enlightening to understand
the structure of Sobolev spaces.}. Then we can define, for any $s\ge 0$, the Sobolev space (see~\cite{adamssobolev}):
\beq \label{eq:defhs}
H^s(\R^N) = \left\{ f \in L^2(\R^N) \left| \int_{\R^N}  |\hat{f}|^2 (\xi) (1+|\xi|^2)^s < \infty \right. \right\}
\eeq
Here $\hat{f}$ is the Fourier transform of $f$.
We cannot expect a probability law (such as a Dirac delta) to belong to some $H^s$ for positive $s$; however, since test
functions in $H^s$ are continuous for $s$ large enough then the Dirac delta
 will 
be in some dual $H^{-s}$ of $H^s$; the dual $H^{-s}$ is a subspace of the space $S'(\R^N)$ of (Schwartz) distributions:
\beq \label{eq:defdualhs}
H^{-s}(\R^N) = \left\{ f \in S'(\R^N) \left| \int_{\R^N}   \frac{|\hat{f}|^2(\xi)}{(1+|\xi|^2)^s} < \infty \right. \right\}.
\eeq
For all $s\in \R$, $H^s$ are Hilbert spaces, the square of the norm of an element $f$ being the integral given in the definition;
$H^0$ reduces to $L^2$.

\subsection{Spaces $\dot{H}^1$ and $\dot{H}^s$} \label{sec:sobolevhdot}

For any connected domain $\Omega$ (in practice for us $\Omega$ is either an open ball in $\R^N$ or the whole space $\R^N$) 
let us introduce the space $BL(L^2)$ of  distributions $f$ such that $\nabla f \in L^2(\Omega)$; we also introduce the 
equivalence relation~: $f \sim g$ if $f=g+c$ where $c$ is a real constant.
Then the quotient of $BL(L^2)$ defined above with respect to this equivalence is denoted by $\dot{H}^1$. One can prove 
as in \cite[Corollary 1.1]{lionshmoins1}
that  $\dot{H}^1$ is a Hilbert space, 
and define its dual $\dot{H}^{-1}$  as in~\cite{rpeyre} which, for our situation, means that
for any two measures $\mu$ and $\nu$ (having the same mass) we obtain a distance $d_{-1}$\footnote{When the boundary $\partial \Omega$ is Lipschitz, replacing the $L^2$ norm in equation \eqref{eq:distancemoins1def}
	 by the $L^\infty$ norm yields instead the 1-Wasserstein distance in the dual form which is used in WGAN.}:
\beq
d_{-1}(\mu,\nu)^2 = \sup \left\{  \int_\Omega f(x) (\mu(dx) - \nu(dx)) \Big| \| \nabla f \|_{L^2} \le 1  \right\}. \label{eq:distancemoins1def}
\eeq

Note that because of the Sobolev embeddings, $\dot{H}^1$ is included in the space of 
continuous functions in dimension $N=1$. \footnote{When there is no ambiguity we denote  $\dot{H}= \dot{H}^1$ .} Thus its dual contains any Dirac mass $\delta_x$, $x\in \Omega$. In particular~: 
\begin{proposition} \label{prop:onedim1h1}
	There exists a universal constant $c_{-1}$ such that for $N=1$ and $x,y \in \Omega$: 
	\beq \label{eq:formula1H1}
	d_{-1}(\delta_x,\delta_y)^2 = c_{-1} |x-y|.
	\eeq
\end{proposition}
\begin{proof} We use the formulation (see~\cite[Section 5.5.2]{book_Filippo}):
$\|\delta_x - \delta_y \|_{\dot{H}^{-1}} = \| \nabla u  \|_{L^2}$ where $u$ is the solution of the Neuman problem
$\partial u / \partial n = 0$ on $\partial \Omega$ and $-\Delta u = \delta_x - \delta_y$ on $\Omega$. But in 1D the solution
is such that $| \nabla u |
={\bf 1}_{[\min(x,y),\max(x,y)]}$.
\end{proof}

It is remarkable that $\dot{H}^{-1}$ is related to the $2$-Wasserstein distance $W_2$ 
through the following relation (see~\cite{rpeyre},\cite[Chap. 7, formula (68)]{ottovillani2000} for details):
if  $\mu$ is a measure with finite second order moment and $\nu$ a small variation, then formally:
\beq \label{eq:relationwassh1}
W_2(\mu,\mu+\epsilon \nu) = |\epsilon| \cdot \| \nu \|_{\dot{H}^{-1}(\mu)} + o(\epsilon),
\eeq
 the norm $\| \cdot \|_{\dot{H}^{-1}(\mu)}$ being the so-called '$\mu$-weighted 
$\dot{H}^{-1}(\mu)$ norm'\cite{adamssobolev}, which belongs to the same family of norms as the 
$\dot{H}^{-1}$ norm that we introduced above (the latter corresponding to $\mu$ being the Lebesgue measure).

In general, one can define for any $s\ge 0$ the homogeneous Sobolev spaces $\dot{H}^s$ as in~\cite{adamssobolev}. In particular functions in spaces with $s > 1/2$ will be  continuous thus their dual will contain all Dirac masses $\delta_x$ and can be used to construct the dual Radon Sobolev distance as described in the next section. For information on how to derive results analogous to the Proposition \ref{prop:onedim1h1} see \cite[equation (1.3)]{brasco2021characterisation}.

%

\subsection{Construction of the dual Radon Sobolev distance} \label{sec:distdef}
We make precise in this section the construction of the Radon Sobolev distance; we recall that this is a version 
of the "sliced" idea used in the literature \cite{CWAE,Deshpande_2018_CVPR,Deshpande_2019_CVPRmax,Kolouri_2018_CVPR,Wu_2019_CVPRgenerative,kolouri2018sliced,cramergan}. 
We suppose thus having at our disposal a distance between two one-dimensional distributions (for instance coming from the
dual of a Hilbert space $\Hcal$ containing continuous functions as in
 the hypotheses of the 
 theorem~\ref{thm:propHilbert}). Note that here $\Hcal$ is a generic name and it has to
be instantiated (for instance $\Hcal= H^1$).

Then the Radon Sobolev distance corresponding to $\Hcal$ is the mean value of the distances of projected distribution 
over all directions on the $N$-dimensional unit sphere $\sphere$; the formal definition is:
\beq
d_{X \Hcal}(\mu,\nu)^2 = \frac{1}{\text{area}(\sphere)} \int_{\sphere}   \|  \theta_*\mu - \theta_*\nu \|_{\Hcal'}^2  d \theta.
\label{eq:formaldefRS}
\eeq
The norm is calculated in the dual space $\Hcal'$ of $\Hcal$, where the measures 
$\theta_*\mu$ and $\theta_*\nu$ belong.
Recall that the projection $ \theta_*\mu$ of the measure $\mu$ on the direction $\theta$ has, when $\mu$ is a sum 
of Dirac masses, the simple expression
$ \theta_* \left( \sum_{k=1}^K \delta_{x_k}  \right) = 
  \sum_{k=1}^K \delta_{\langle x_k, \theta \rangle}$.
\begin{remark} 
When  
$ \| \delta_x - \delta_y \|_{\Hcal'}$ only depends on $|x-y|$ (as is the case for translation invariant norms
and in particular for the Sobolev spaces in sections~\ref{sec:sobolevhs} and ~\ref{sec:sobolevhdot}) then
there exists a function $g$ such that for any
$x,y \in \Omega$:
$d_{X \Hcal}(\delta_x,\delta_y)^2 = g(| x - y |)$. In this case 
all that is required to compute the distance 
$d_{X \Hcal}$ is the knowledge of a real function $g$ of one variable.
For instance the formula in equation~\eqref{eq:distsumdiracs} 
becomes:
\begin{multline}
d_{X \Hcal} \left( \frac{1}{K} \sum_{k=1}^K \delta_{x_k}, \frac{1}{J} \sum_{j=1}^J \delta_{y_j} \right)^2
=  \sum_{k=1}^K \sum_{j=1}^J  g( | {x_k}- {y_j} |) 
\\
- \frac{\sum_{k=1}^K\sum_{k'=1}^K  g(| {x_k}- {x_{k'}} |)}{2 K^2}  
-\frac{\sum_{j=1}^J\sum_{j'=1}^J  g( | {y_j}- {y_{j'}} |)}{2 J^2}.	
\label{eq:distsumdiracs2}
\end{multline}
	Note that the translation invariance, i.e., that $ \| \delta_x - \delta_y \|_{\Hcal'}$ only depends on $|x-y|$, is not a consequence of previous considerations; the meaning of this remark is to argue that such a situation is  favorable and worth investigating.
 \label{rem:simpledistance}
\end{remark}
\begin{remark}
	An anonymous referee pointed out to us that this procedure can be
	 extended to averaging over projections onto $d$-dimensional subspaces (for $1\le d \le N$); 
	 this enlarges the choice of Radon-Sobolev norms and has the potential to 
	 improve the outcomes.
\end{remark}
\subsection{Explicit formulas for the distance}
With these provisions we can prove the following
\begin{theorem} \label{thm:propHilbert}
	Let $M>0 $ (M can be +$\infty$) and $\Omega$ the ball of radius $M$ in $\R^N$ (whole $\R^N$ if $M=\infty$).
	Let  $(\Hcal,\| \cdot \|_{\Hcal})$ be a Hilbert space of (classes of) real functions of one variable (such as $H^s(\R)$ or $\dot{H}^s(\R)$)
	on the domain $\mathopen]-M,M\mathclose[$ such that 
	$\Hcal$ is included in the set of continuous functions $C^0(\Omega)$.
	Denote by $(\Hcal',\| \cdot \|_{\Hcal'})$ the dual of $H$, $\probalaws$ the set of probability laws on $\Omega$ and
	$d_{X \Hcal}$ the distance on $\probalaws$ associated to $\|\cdot \|_{\Hcal'}$ as described in section~\ref{sec:distdef}.
	Then  $\probalaws$ can be seen as an affine subset of some Hilbert space with the distance given by $d_{X\Hcal}$ i.e., can be isometrically embedded into a Hilbert space.
\end{theorem}
\begin{proof}
The distance $d_{X\Hcal}$ is constructed following the procedure described
in section~\ref{sec:distdef};
 it builds on
 the space of square integrable functions from $\sphere$ to $\Hcal'$: to each distribution
$\mu$ we associate the function $f_\mu : \sphere \to \Hcal'$ by the relation
$\theta \in  \sphere \mapsto f_\mu(\theta)= \theta_* \mu \in \Hcal'$. 
This association is injective.
Since the dual $\Hcal'$ of $\Hcal$ is also a Hilbert space,
the set of such functions $\{f_\mu; \mu \in \probalaws \}$ is a subset of 
$L^2(\sphere,\Hcal')$ which is a Hilbert space with the usual scalar product
$\langle g_1,g_2 \rangle = 
 \frac{1}{\text{area}(\sphere)} 
\int_{\sphere} \langle g_1(\theta), g_2(\theta)\rangle_{\Hcal'} 
d \theta$.
Thus to each element of  $\mu$ in the metric space $(\probalaws,d_{X\Hcal})$ one can associate an element 
${\mathcal F}(\mu)= f_\mu \in L^2(\sphere,\Hcal')$; by the very definition of $d_{X\Hcal}$ the mapping 
 ${\mathcal F}$ conserves the distance, and the image is a subset of the Hilbert space $L^2(\sphere,\Hcal')$ (considered with its scalar-product induced distance); we have established thus the embedding mentioned by the conclusion.
\end{proof}

\begin{corrolary} \label{cor:propdist}
Under the assumptions of the theorem \ref{thm:propHilbert} the distance 
 $d_{X\Hcal}$ is such that:
\begin{enumerate}
	\item 
 any line $\mu_t = (1-t)\mu_0 + t \mu_1  \subset \probalaws$, $t\in [0,1]$ is a geodesic and 
$d_{X\Hcal}$ is convex on $\mu_t$;
	\item 
 for any $\nu \in \probalaws$ the distance $d_{X \Hcal}$ satisfies~\eqref{eq:ddd_id};
	\item 
 the distance $d_{X \Hcal}$ satisfies relation~\eqref{eq:distsumdiracs}.
\end{enumerate}

\noindent In particular, up to a multiplicative constant, for $\Hcal=\dot{H}^{1}$:
\begin{multline}
d_{X \dot{H}^{1}}(\mu,\nu)^2= \E_{X \sim \mu, Y \sim \nu, X \indep Y  } \|X - Y \| 
\\
-\frac{	\E_{X,X' \sim \mu, X \indep X'  } \|X - X' \| + \E_{Y,Y' \sim \nu, Y \indep Y'  } \|Y - Y' \|}{2}, 
 \label{eq:energynorm}
\end{multline}
where the indices in expectation symbols mean that $X$, $X'$, $Y$, $Y'$ are independent random variables, the first two with distribution $\mu$ and the last two with distribution
 $\nu$. 
\end{corrolary}
\noindent {\bf Proof:} To prove the first two points it is enough to recall that property~\eqref{eq:ddd_id} is true in a Hilbert space because there the distance (squared) is the norm (squared)
and the identity follows from the expansion using the scalar product. 

The third point follows as in Proposition~\ref{prop:computation}.

\begin{remark}
	Under the hypothesis of the remark \ref{rem:simpledistance}
	the procedure in section \ref{sec:distdef} allows to resume the 
	computation of any  
	distance $d_{X \Hcal}(\frac{1}{J} \sum_{j=1}^J \delta_{x_j}, \normal{0}{I_N})$	
	to two functions 
	$\gxy$ and $\gxn$ such that
	\beq \label{eq:defgxy}
	d_{X \Hcal}(\delta_x, \delta_y)^2 = \gxy (\| x-y\|), \ \forall x,y \in \Omega,
	\eeq
	\beq \label{eq:defgxn}
	d_{X \Hcal}(\delta_x, \normal{0}{I_N} )^2 = \gxn (\| x\|), \ \forall x \in \Omega,
	\eeq
 where $\normal{0}{I_k}$ designates the standard multivariate $k$-dimensional normal ($k\ge 1$). 
 This is an example of "non-local computability"  invoked in section \ref{sec:nonlocalcalculability}.
	These functions can be used in the RS-Variational Auto-Encoder as detailed in section~\ref{sec:xsvae}. 
	Note that,  contrary e.g., to sliced Wasserstein implementations, here there is no
	need to discretize the directions of the $N$-dimensional sphere $\sphere$.
	The precise formulas and approximation of  
	$ \gxy$ and $ \gxn$ when  $\Hcal=\dot{H}^1$ are the object of the 
	Proposition  	\ref{prop:formulash1}
	whose proof is given in	\ref{sec:analytic}.
	 \label{rem:defgxygxn}
\end{remark}


\begin{proposition} 
For $\Hcal=\dot{H}^1$, up to multiplicative constants,
$\gxy(z)=|z|$ and 
$\gxn(z) = c_{N0}+ \sqrt{z^2 + c_{N1}} +  O \left(
| z|^4 \right)$ (in the limit $z\to 0$)  . More precisely:
\begin{enumerate}
\item For any $x,y \in \Omega$:
$d_{X\Hcal}(\delta_x,\delta_{y})^2 = | x-y|$ (up to a multiplicative constant taken as $1$);
\item Define for $X=(X_1,...,X_N) \sim \normal{0}{I_N}$:
\begin{equation}
	d_N(a)= \E [|(a,0, ..., 0) - X|]=\E\left[\sqrt{(a-X_1)^2 + X_2 + ... + X_N^2} \, \right].
	\label{eq:definitiondn}
\end{equation}
Then
\begin{equation}
d_{X \Hcal}(\delta_x,\normal{0}{I_N})^2= \gxn(|x|)= \xi_N( |x |),
\label{eq:defxi}
\end{equation}
where
\begin{align}
\xi_N(a) &= d_N(a) - \frac{d_N(0)}{\sqrt{2}} \nonumber\\&= 
\sum_{k=0}^\infty \frac{ (a^2/2)^k e^{-a^2/2}}{k!} 
\sqrt{2} \frac{\Gamma\left( \frac{N+1}{2} +k\right)}{\Gamma\left( \frac{N}{2} +k\right)}
- \frac{\Gamma\left( \frac{N+1}{2} \right)}{\Gamma\left( \frac{N}{2}\right)}.
	\label{eq:xi_as_dn}
\end{align}
Here $\Gamma$ is the Euler gamma function (for instance $\Gamma(n+1) = n!$ for any integer $n$).
\item We also have
\begin{equation}
\frac{d}{da}\xi_N(a)  = a \left(  d_{N+2}(a)-d_N(a)  \right), \ \forall a \in \R.
\label{eq:dxida}
\end{equation}
\item In particular: 
\beqn 
& \ & d_{X\Hcal}(\delta_x,\normal{0}{I_N})^2 = 
c_{N0}+ \sqrt{|x|^2 + c_{N1}}
+  O \left(
|x|^4 \right).
\label{eq:normalh1formulax}
\eeqn
Thus the following approximation is exact 
up to  
$|x |^4$:
\begin{multline}
d_{X \Hcal} \left( \frac{1}{K}\sum_{k=1}^K \delta_{x_k},\normal{0}{I_N} \right)^2 \simeq
c_{N0}
\\
+ \frac{\sum_{k=1}^K \sqrt{| x_k |^2 + c_{N1}} }{K} 
- \frac{\sum_{k=1}^K\sum_{k'=1}^K \!\!   | {x_k}- {x_{k'}} |}{2 K^2},
\label{eq:normalh1formula}
\end{multline}
where $c_{N0}, c_{N1}$ are defined in \eqref{eq:defcn01}.
\end{enumerate}
\label{prop:formulash1}
\end{proposition}

\begin{remark}
	If for some $\Hcal$ and some constant $\beta \in ]0,2[$ we have $\gxy(z)=|z|^\beta$ $\forall z$, a formula analogous to \eqref{eq:xi_as_dn} can be proven using the same techniques:
\begin{align}
	\gxn(a) &= 
	\sum_{k=0}^\infty \frac{ (a^2/2)^k e^{-a^2/2}}{k!} 
	\sqrt{2}^\beta \frac{\Gamma\left( \frac{N+\beta}{2} +k\right)}{\Gamma\left( \frac{N}{2} +k\right)}
	- 2^{\beta-1}\frac{\Gamma\left( \frac{N+\beta}{2} \right)}{\Gamma\left( \frac{N}{2}\right)}.
	\label{eq:gn_for_xbeta}
\end{align}
\end{remark}

\section{RS-VAE: Radon Sobolev Variational Auto-Encoder} \label{sec:xsvae}

\subsection{Algorithm}

Based on the previous results, we propose a new type of VAE using the above distances. 
Compared with a GAN
the VAE has the advantage of using a fixed reference latent distribution and does not need to
look for a suitable "feature space" to express the distance in. This means that, as in~\cite{CWAE}, 
one can pre-compute the 
distance from a Dirac mass to a target latent distribution (here a standard multivariate normal) which speeds up the computations even more.

 In order to compare with results from the 
literature we use the space $\Hcal=\dot{H}$ that has been studied before and that has also interesting properties, cf. 
equation~\eqref{eq:relationwassh1}.

Let us fix the following notations: $\theta_e$ are the parameters of the encoder network
and $\theta_d$ the parameters of the decoder. The encoding is a (parametrized ) function transforming some 
real sample $X \in \R^I$ to the latent space $\R^N$, i.e. $E_{\theta_e}(X) \in \R^N$. 
The goal of this part is to have the distribution $\preal$ transported to the latent distribution (standard multivariate normal)
on the latent space; the corresponding part of the loss functional is
$d_{X \dot{H}^1}((E_{\theta_e})_* \preal, \normal{0}{I_N})^2$.
The decoding part takes $z \in \R^N$ and through the decoding function $D_{\theta_d}$ maps it to the 
initial space $\R^I$; the goal is to provide accurate reconstruction of the real samples,
 which corresponds to the minimization of the mean squared error
$\E_{X \sim \preal} \| [D_{\theta_d} \circ E_{\theta_e}]_* (X) -  X\|^2$.
We obtain the procedure described in Algorithm \ref{alg:xsvar}.
 As a side remark note that in practice the sampling step "sample $X_1,...,X_k \sim \preal$ (i.i.d)"
	of	Algorithm \ref{alg:xsvar} is to be understood in the usual sense of using batches and epochs to draw from the empirical data set
	$\{Y_\ell; \ell=1, ...,L\}$.

\begin{algorithm}
	\caption{Radon-Sobolev Variational Auto-Encoder (RS-VAE)}
	\label{alg:xsvar}
	\begin{algorithmic}[1]
		\Procedure{RS-VAE}{}
		\State $\bullet$ set {batch size $K$, latent dimension $N$, $\lambda \ge 0$}
		\State $\bullet$ compute constants $c_{N0}$, $c_{N1}$ from equation~\eqref{eq:defcn01}
		\While{(stopping not reached)}
		\State $\bullet$ sample $X_1,...,X_K \sim \preal$ (i.i.d).
		\State $\bullet$ propagate the real sample through the encoding network;
		\State $\bullet$ compute the latent loss: 
		\begin{multline}
	\mathit{Loss}_{\mathit{lat}}(\theta_e,\theta_d) :=c_{N0} +\frac{\sum_{k=1}^K \sqrt{| E_{\theta_e} (X_k) |^2 + c_{N1}} }{K}
	 \\ 
	 -\frac{\sum_{k=1}^K\sum_{k'=1}^K \!\!   | E_{\theta_e} (X_k)- E_{\theta_e} (X_{k'}) |}{2 K^2}.
 \end{multline}
		\State $\bullet$ propagate the real sample through the decoding network;  
		\State $\bullet$ compute the reconstruction loss: 
		
		$\mathit{Loss}_{\mathit{rec}}(\theta_e,\theta_d) :=\frac{1}{K}\sum_{k=1}^K \| D_{\theta_d}(E_{\theta_e}(X_k)) - X_k\|^2$
		\State $\bullet$ compute the global loss: 

		$\mathit{Loss}_{\mathit{global}}(\theta_e,\theta_d):=\mathit{Loss}_{\mathit{rec}}(\theta_e,\theta_d)  + \lambda \mathit{Loss}_{\mathit{lat}}(\theta_e,\theta_d)$
		\State $\bullet$ backpropagate $\mathit{Loss}_{\mathit{global}}(\theta_e,\theta_d)$ and update
		 parameters $\theta_e,\theta_d$ (a minimization step).		 
		\EndWhile\label{euclidendwhile}
		\EndProcedure
	\end{algorithmic}
\end{algorithm}

\subsection{Numerical results}

We tested the RS-VAE on several datasets from the literature. The goal here is to show that this procedure is comparable with 
other probability distances used in generative networks while still benefiting from a fast computation and no hyperparameters to choose (except latent space size
and batch size that all VAEs deal with).

The code is available in the supplementary material and 
 as a Github repository;\footnote{Available at \url{https://github.com/gabriel-turinici/radon_sobolev_vae}.}
the comparison is made with the algorithm from~\cite{CWAE}.
The datasets are MNIST, Fashion-MNIST and CIFAR10 and the encoder-decoder  architecture is taken from the reference (and recalled in \ref{sec:architecture}).

In all cases we used the additive version of the loss functional because the "log" introduced in \cite{CWAE} 
for the distance term would 
prevent from having a convex loss (the "log" being concave). In practice the scaling constant $\lambda$ can either be chosen
by trial and error or by running first two optimizations, one with only reconstruction loss and the other with only  
latent loss, and taking the quotient of the standard deviations in the oscillations seen in the result. In practice
it has been set to $\lambda=100$ for all tests.

\subsubsection{Relevance of both loss terms} \label{sec:num2terms}

We first test whether in the RS-VAE  both loss terms are effective. We only show the results for MNIST dataset
but the conclusions are similar for all datasets. Namely we consider three runs:

\noindent $\bullet$ optimization of the reconstruction loss only ($\lambda=0$);

\noindent $\bullet$ optimization of the latent loss only (equivalent to taking $\lambda \to \infty$);

\noindent $\bullet$ full optimization ($\lambda=100$).

\begin{figure}
\hspace{2.5cm} 	Samples \hspace{4.cm} Reconstructions 

	\includegraphics[width=.495\linewidth]{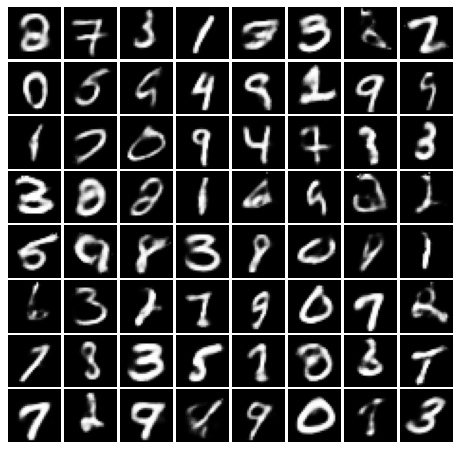}
	\includegraphics[width=.495\linewidth]{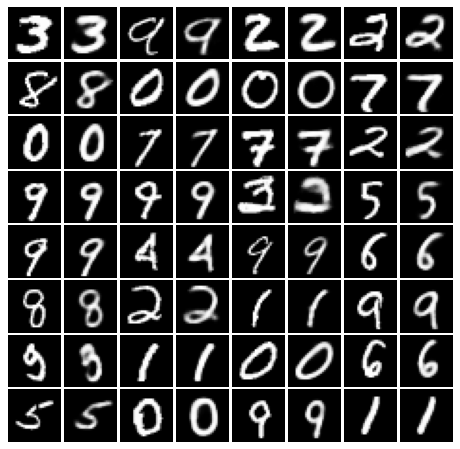}

	\includegraphics[width=.495\linewidth]{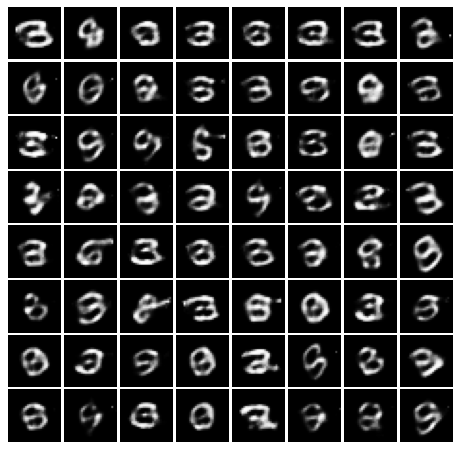}
	\includegraphics[width=.495\linewidth]{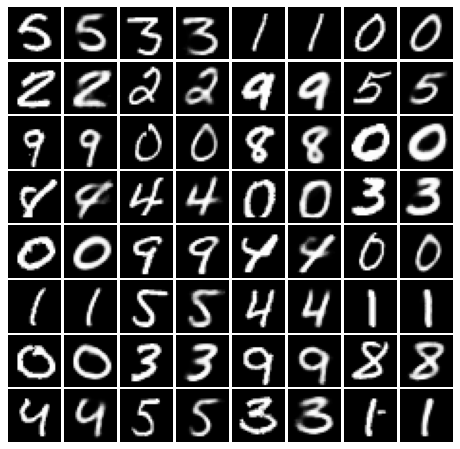}
	\caption{MNIST dataset.
		{\bf Top Left:}  generated samples (full loss functional).
		{\bf Top Right:} reconstruction quality (full loss functional).
		{\bf Bottom Left:}  generated samples (reconstruction only loss functional).
	{\bf Bottom Right:} reconstruction quality (reconstruction only loss functional).
}	\label{fig:resultsMNIST-short}
\end{figure}

The results for the latent loss only are given in figure~\ref{fig:consistency}; as expected the network  
	 generates "white noise" images.
The results for the reconstruction loss only and full optimization are given in figure \ref{fig:resultsMNIST-short}. The first situation results in good reconstruction but poor 
generation; the advantage of having both terms is obvious from the figure.

When only optimizing the reconstruction loss the reconstruction error drops from
$39.5$  (epoch $1$) to around $11.5$  (epoch $100$).  
 To appreciate these values, let us mention that the mean error of a trivial reconstruction 
 procedure that  
assigns, to some input object $Y_k$,
 a random object $Y_{k'}$ from the database, is given by the formula   
$\frac{\sum_{k,k'=1}^K \| Y_k - Y_{k'}\|^2}{K^2}$ and 
is worth $105.70$ for our dataset\footnote{Same value equals 
$135.84$ for FMNIST and $379.50$ for CIFAR10.}. 
So, the reconstruction error can be considered
 to be roughly $10$ times better than a random one, which is a fairly good result
 as one can observe readily on figure \ref{fig:resultsMNIST-short}.
The Radon Sobolev distance oscillates at  around $1.0$; this is a quite bad value, as expected: indeed, the average value of the loss function for a Dirac mass drawn according
to the $\normal{0}{I_N}$ law, that is 
$\E_{x \sim \normal{0}{I_N}}[d_{XH}(\delta_x,\normal{0}{I_N})^2]$ is worth $3.88$\footnote{The value is the same for FMNIST and is $11.26$ for CIFAR10.}: so, here
the latent loss is of the same order of magnitude as if all the inputs had been 
concentrated at the same point of the latent space.
 
 When optimizing the latent part only the reconstruction loss
remains at high values (depending on the run between $40$ and $180$) and the 
latent distance reaches small values (around $10^{-4}$ and below).
When using both terms, the reconstruction error drops down to around $11.5$ and 
the distance around $5 \times 10^{-3}$, depending on the run.

This behavior remains visible in all tests we performed.

\subsubsection{Comparison with other "sliced" distances} \label{sec:comparison}

One of the main interests of the distance that we introduce is to be able to provide a computational efficient alternative to previous metrics such as the sliced Wasserstein~\cite{Kolouri_2018_CVPR,Deshpande_2018_CVPR,kolouri2018sliced,Wu_2019_CVPRgenerative,Deshpande_2019_CVPRmax} 
(referenced as "SW" in the following) 
and the Cram\'er-Wold distances~\cite{CWAE} (named "CW").
As such it is interesting to know whether optimizing the Radon Sobolev distance (referenced as "RS") is a 
proxy for minimizing the SW or CW distances. In fact,  
evidence is already present in the literature that optimizing the CW distance 
and SW distances is similar; given this context, we will only compare with one distance on each dataset. 

\begin{figure}
\includegraphics[width=.45\linewidth]{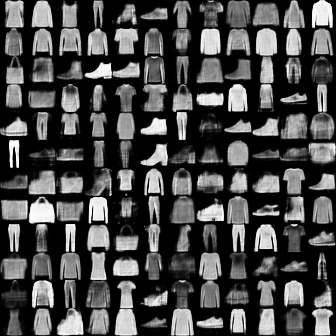}
	\includegraphics[width=.45\linewidth]{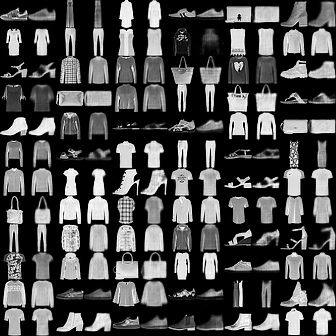}
	\caption{  Fashion-MNIST dataset.
		{\bf Left:}  generated samples.
		{\bf Right:} reconstruction quality.
}	\label{fig:resultsFMNIST}
\end{figure}

\begin{figure}
\begin{center}
\includegraphics[width=.49\linewidth]{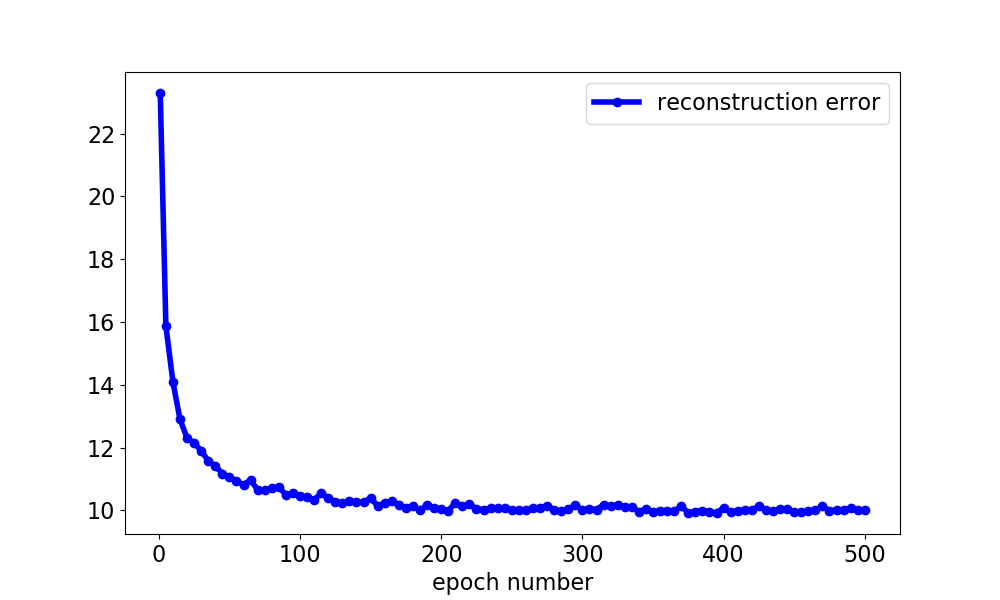}
\includegraphics[width=.49\linewidth]{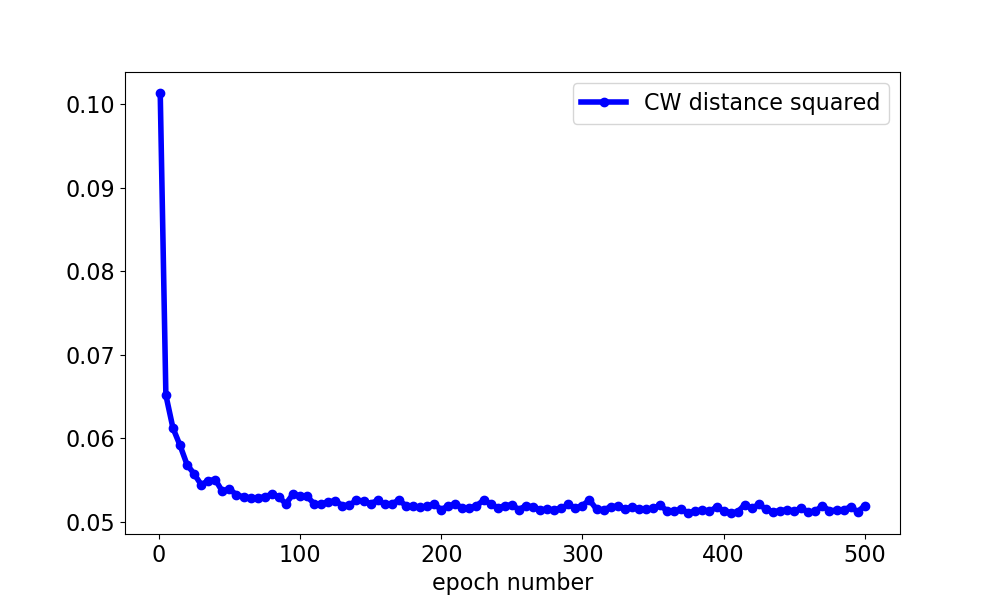}
\includegraphics[width=.5\linewidth]{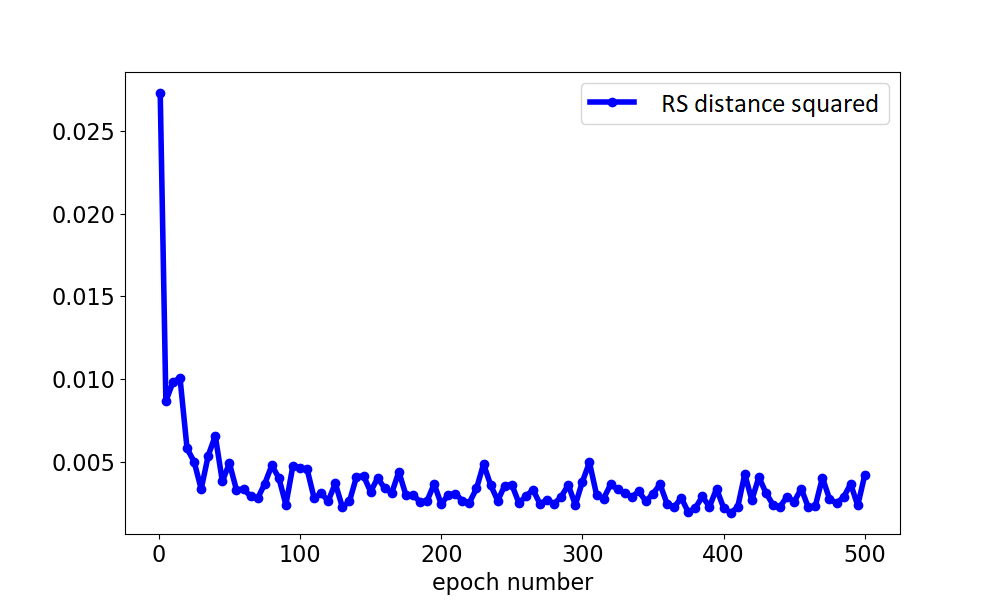}
\end{center}	
	\caption{Convergence of the RS-VAE procedure: plot of the reconstruction error, RS and CW distances for the Fashion-MNIST dataset.
		RS and CW distances move in similar directions (Pearson' R correlation coefficient from epoch $100$ to epoch $500$ equals 
			$83\%$)
		a good indication that they are equivalent for practical purposes. 
		 See also 
		figure \ref{fig:scatter} for a scatter plot.
}
	\label{fig:FMcomparaisonCW}
\end{figure}

A comparison between the RS and CW distances on the Fashion-MNIST dataset 
is presented in Figures~\ref{fig:FMcomparaisonCW} (see Figure~\ref{fig:resultsFMNIST} for illustrative results on Fashion-MNIST). 

\begin{figure}
\begin{center}
\includegraphics[width=0.75\linewidth]{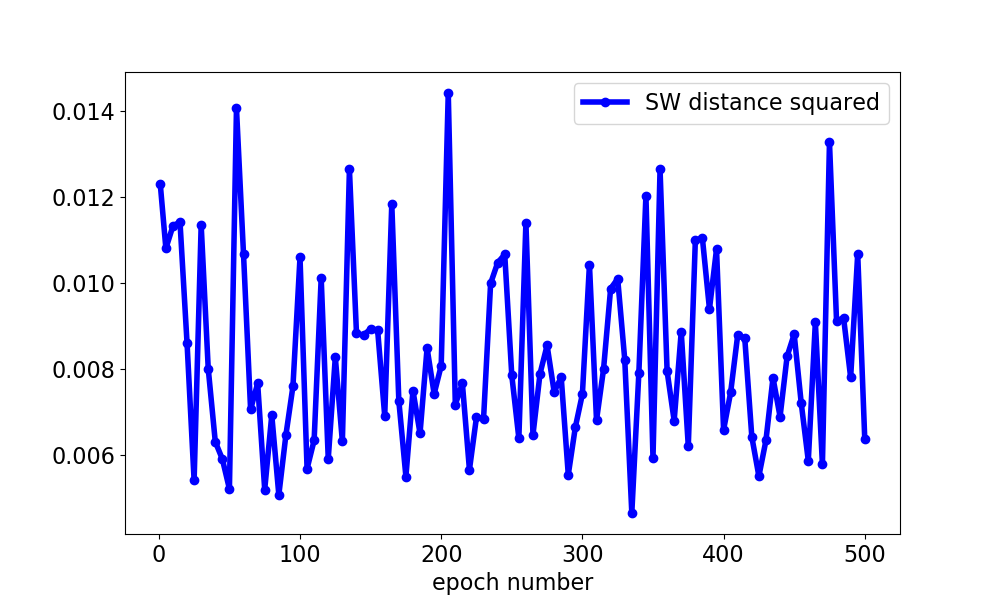}
\includegraphics[width=0.75\linewidth]{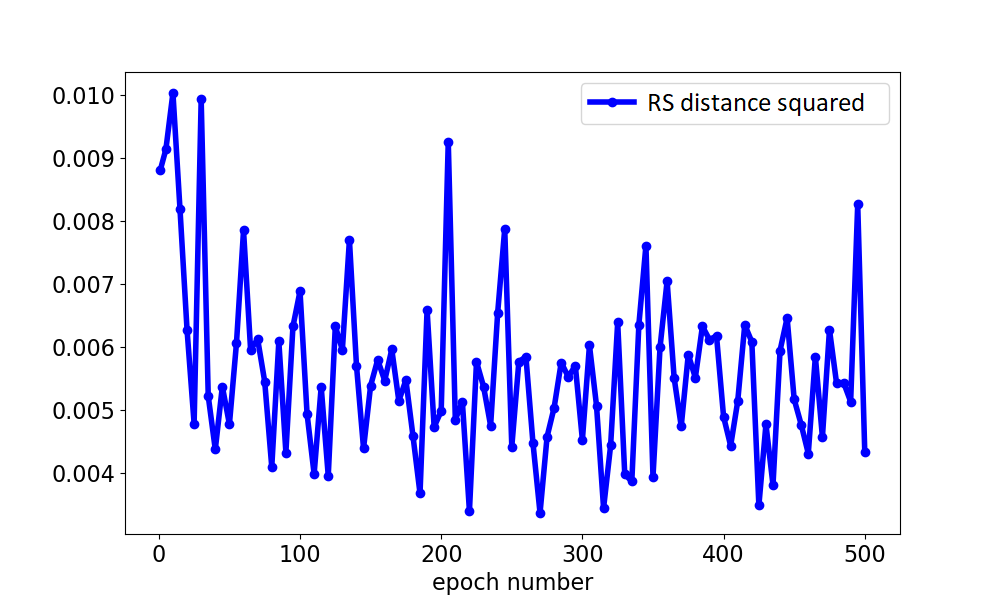}
\end{center}
\caption{Convergence of the RS-VAE procedure: comparison of the SW and RS distances for the MNIST dataset.
	Due to convexity, the convergence of  the distance
is very fast and we see only the  oscillation around the correct value. However this is interesting exactly because
it allows to see that both distances move in a correlated manner (Pearson' R correlation coefficient from epoch $100$ to epoch $500$ equals $64\%$),
 indicating that they are very similar for practical purposes. Thus one can use RS distance
instead of SW. Note that we do not expect a perfect linear relationship. See also 
figure \ref{fig:scatter} for a scatter plot.
}
\label{fig:M_comparaisonSW}
\end{figure}

\begin{figure}
	\begin{center}
		\includegraphics[width=0.45\linewidth]{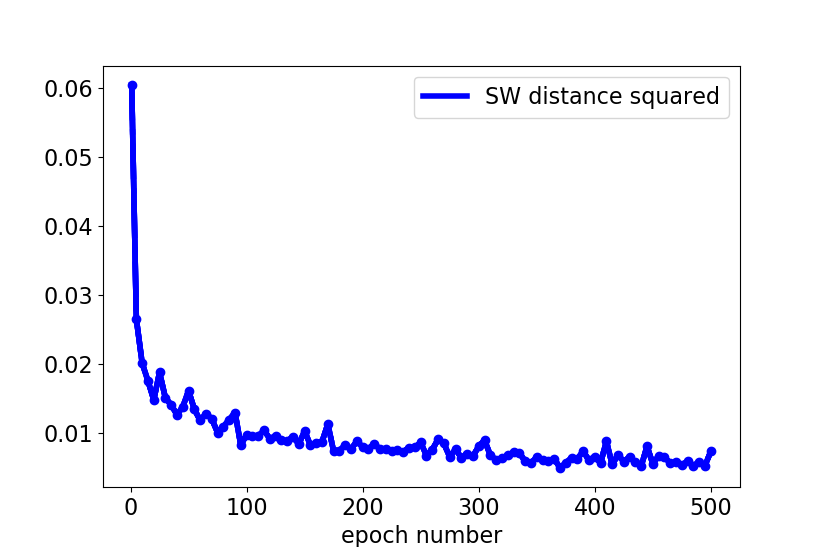}
		\includegraphics[width=0.45\linewidth]{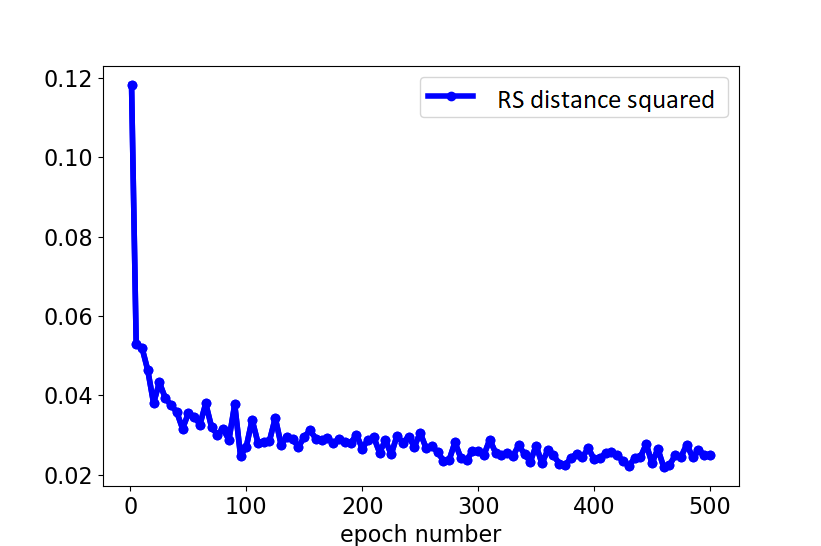}
	\end{center}
	\caption{Convergence of the RS-VAE procedure: comparison of the SW and RS distances for the CIFAR10 dataset. Compared with
	MNIST dataset in Figure~\ref{fig:M_comparaisonSW} the convergence is slower. 
The Pearson's R coefficient from epoch $100$ to epoch $500$ equals $67\%$
 See also figure \ref{fig:scatter} for a scatter plot.
}.
	\label{fig:C10_comparaisonSW}
\end{figure}

\begin{figure}
\begin{center}
	\includegraphics[width=.49\linewidth]{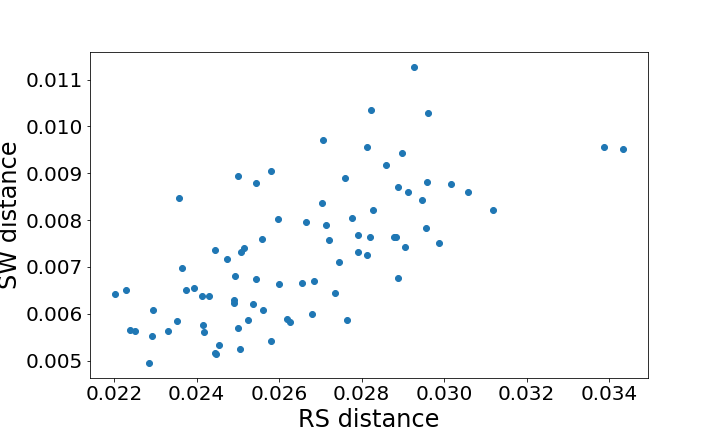}
	\includegraphics[width=.49\linewidth]{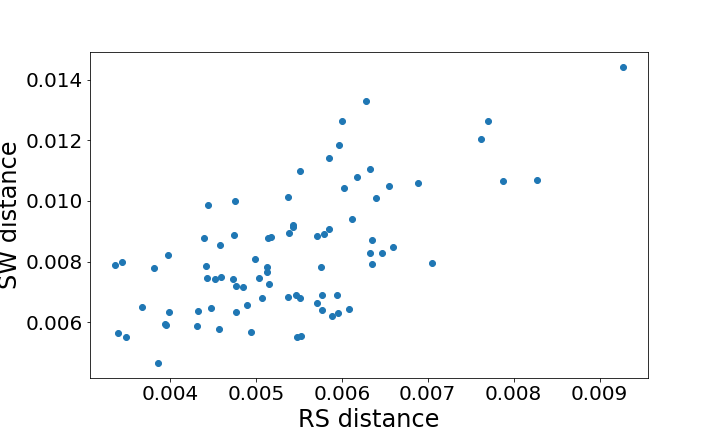}
	\includegraphics[width=.5\linewidth]{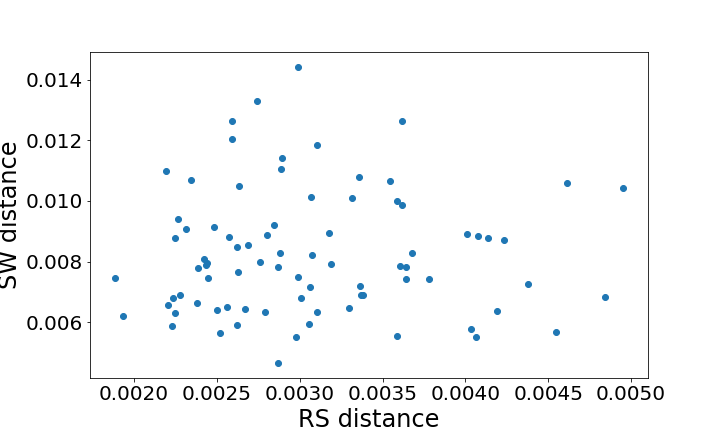}
\end{center}	
\caption{Scatter plot of the RS versus SW distance (top left: CIFAR10, top right: MNIST, bottom:  FMNIST) from epoch $100$.
}
\label{fig:scatter}
\end{figure}

Comparisons between the SW and RS distances on MNIST and CIFAR10 are presented in 
Figures~\ref{fig:M_comparaisonSW}, \ref{fig:C10_comparaisonSW} and \ref{fig:scatter}. Both show that the RS distance is a practical proxy
for the SW and CW distances.

\begin{figure}
	\includegraphics[width=.45\linewidth]{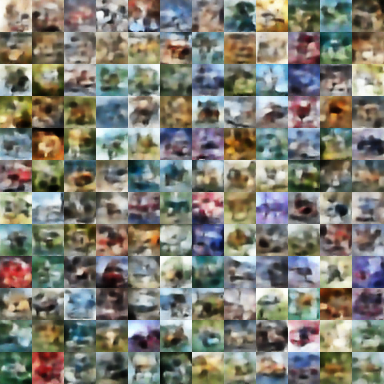}
	\includegraphics[width=.45\linewidth]{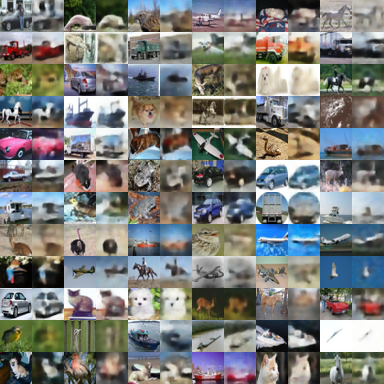}
	\caption{CIFAR10 dataset.
		{\bf Left:}  generated samples. The network architecture unfavorably influences the reconstruction quality and the sampling quality.
		 This is a known behavior for the CIFAR10 dataset which is both low resolution and very diverse as compared to other benchmarks (MNIST, FMNIST, even CelebA, see \cite{CWAE}). Nevertheless the generator is able to synthesize correctly structured backgrounds and plausible forms. 
		{\bf Right:} reconstruction quality.
	}	\label{fig:resultsCIFAR10}
\end{figure}

Finally, we present 
in table \ref{table:1} a more quantitative validation of the generative quality.
	For each dataset we consider three runs obtained by setting the latent loss driven by the RS, SW and CW distances respectively 
	(the reconstruction loss remains the same). 
	The metric used to quantify the quality of the generation is the 
	'Fr\'echet Inception Distance'\cite{fid_metric} which is the state of the art evaluator metric for image generators.	
 The RS distance obtains similar results to all other metrics discussed in this section 
(but has the advantages discussed in section \ref{sec:literature}); 
 more specifically, 
\begin{itemize}
\item 	the  SW distance requires integration over all possible directions, 
similar to our formula \eqref{eq:formaldefRS}; for SW the integration is 
performed numerically by sampling over all points $\theta$ on the 
$N$-dimensional unit sphere, while for us the distance can be resumed to two functions (only depending on the dimension of the latent space) that can be precomputed, tabulated or approximated. This avoids a Monte Carlo approximation of a high-dimensional integral arising at each optimization step.
	 
\item the CW distance uses a hyper-parameter to control the smoothing: to compute the metric, the input measures are smoothed and the $L^2$ distance between their densities computed; although heuristics allow to have some indications on the optimal choice of the parameter, exceptions always appear and thus affect the quality of the numerical result.
\end{itemize}


\begin{table}[htb]
\begin{tabular}{c|ccc}
{\scriptsize	
\backslashbox{Dataset}{Distance} }  & Radon Sobolev & Sliced Wasserstein & Cramer-Wold \\ \hline 
MNIST   & 29   & 30 (vs. 25-30) & 28 (vs. 24)\\
FMNIST  & 59   & 59  (vs. 62-74) & 53 (vs. 57)\\
CIFAR10 & 117 & 120  (vs. 120-142) & 118 (vs. 120)\\
\end{tabular}
\caption{We measure the generation quality with the FID index. 
	In numerical tests we used the same parameters as those  described previously (including the number of epochs). 
	The FID metric is computed with $10\ \ \!\! 000$ images from
the validation set. The values cited in \cite{CWAE} are recalled in parenthesis. As for the convergence, the FID metric is 
very similar among all three drivers : CW, RS, SW.} \label{table:1}
\end{table}

\section{Conclusion}

We introduced in this work a class of probability distances to be used in generative modeling; all members of the class share the important properties
of convexity and fast evaluation which were not always present in previous works (Wasserstein, sliced distances, ...).
Each distance corresponds to a Hilbert space of functions (Sobolev spaces in this work) and is constructed using the Radon transform. 

To illustrate the effectiveness of the metrics, we considered the particular case of the Sobolev space $\dot{H}^{-1}$ which
gives a distance already present in the literature, for which we derive and use novel fast evaluation formulas.

The resulting procedure, called RS-VAE (to recall the Radon Sobolev construction) is shown to perform well on
standard datasets.

\section{Acknowledgments}
We want to thank  two anonymous referees for helpful suggestions and comments.

\newpage 

\appendix

\section{Analytic computations of $\gxy$ and $\gxn$} \label{sec:analytic}

We recall that $\gxy$ is the square of distance between two Dirac masses (the argument is the Euclidian distance between the masses) 
	and $\gxn$ is the distance (squared) between a Dirac mass and the multivariate normal distribution.

We give first a result used in the proof of the proposition \ref{prop:formulash1}. Recall that the noncentral $\chi$ distribution with $k$ degrees of freedom and noncentrality parameter $\lambda$ is the law of the 
	random variable 
	$\sqrt{\sum_{i=1}^k (\lambda_i +X_i)^2}$ where 
$(X_1,...,X_k) \sim \normal{0}{I_k}$, and 
$\lambda=\sqrt{\sum_{i=1}^k \lambda_i^2}$. We denote by 
$\chi_k(\lambda)$ the law of such a variable. When $\lambda=0$ we obtain in particular 
a central $\chi$ variable, whose law is denoted $\chi_k$.	We will see that a noncentral $\chi$ variable is a Poisson weighted mixture of central $\chi$ variables.
	
\begin{lemma}
\begin{enumerate}
\item Let $\ell \ge 1$, $Z_i \sim \chi_{\ell+2i}$, $i \ge 0$ and $Y$ be a Poisson random variable of mean $\lambda^2/2$ independent of all $(Z_i)_{i\ge 0}$. Then:
\begin{equation}
U = \sum_{i=0}^\infty \mathbbm{1}_{Y=i} \cdot Z_i, 
\label{eq:noncentralchi}
\end{equation}	
 is a noncentral $\chi$ variable with $\ell$ degrees of freedom and noncentrality parameter $\lambda$.
\item In particular
\begin{equation}
\E[U] = \sum_{i=0}^\infty \frac{ (\lambda^2/2)^i e^{-\lambda^2/2}}{i!} 
\sqrt{2} \frac{\Gamma\left( \frac{\ell+1}{2} +i\right)}{\Gamma\left( \frac{\ell}{2} +i\right)}.
\label{eq:formulanoncentralchimean}
\end{equation}	
\item As a consequence the function $d_N$ defined in \eqref{eq:definitiondn} can be written as:
\begin{equation}
	d_N(a)=\sum_{k=0}^\infty \frac{ (a^2/2)^k e^{-a^2/2}}{k!} 
	\sqrt{2} \frac{\Gamma\left( \frac{N+1}{2} +k\right)}{\Gamma\left( \frac{N}{2} +k\right)}.
	\label{eq:dnformula}
\end{equation}
\end{enumerate}
\end{lemma}
\begin{proof}
	We recall the classical interpretation of a noncentral $\chi^2$ distribution as a Poisson mixture of central
	$\chi^2$ distributed variables~\cite{noncentralchi,johnson_kotz_stat_book}; by taking the square root one obtains \eqref{eq:noncentralchi}.
To obtain \eqref{eq:formulanoncentralchimean} it is enough to recall that 
the mean of a central $\chi$ variable with $m$ degrees of freedom) equals
$\sqrt{2} \frac{\Gamma((m+1)/2)}{\Gamma(m/2)}$ and that
$\P[Y=i] =\frac{ (\lambda^2/2)^i e^{-\lambda^2/2}}{i!}$; finally, \eqref{eq:dnformula}
is a simple consequence of \eqref{eq:definitiondn},\eqref{eq:formulanoncentralchimean} and the definition of a
noncentral $\chi$ variable.
\end{proof}

{\it Proof of the proposition \ref{prop:formulash1}} 
Take $x,y \in \Omega$ and $Z, Z'$ two independent random variables that follow each a $N$-dimensional standard normal distribution.
 When  $\Hcal=\dot{H}^1$ we already saw in equation~\eqref{eq:formula1H1}
that the distance is translation invariant in one dimension i.e., depending only on 
 $| \langle x-y, \theta\rangle |$.
Then, by symmetry, the mean over all directions $\theta \in \sphere$ of
 $| \langle x-y, \theta\rangle |$ is a multiple of $\|x-y\|$ (see also~\cite{SZEKELY13} which introduces 
 the same distance from another point of view).
 
We use now \eqref{eq:energynorm} with $Z,Z' \sim \normal{0}{I_N}$: 
\begin{equation}
d_{X \Hcal}(\delta_x,\normal{0}{I_N})^2= \E |x - Z | -\frac{	\E |Z - Z' |}{2} =  \E |x - Z | - \frac{\sqrt{2}}{2} \E |Z|.
\label{eq:formulainproof_xi_as_dn}
\end{equation}
 Note that, given the definition of $d_N$, the relation \eqref{eq:xi_as_dn}  is but a re-writing of \eqref{eq:formulainproof_xi_as_dn}.
 Equation \eqref{eq:dxida} follows from \eqref{eq:xi_as_dn} and \eqref{eq:dnformula} by derivation with respect to $a$.
 
\begin{figure}
\includegraphics[width=0.96\linewidth]{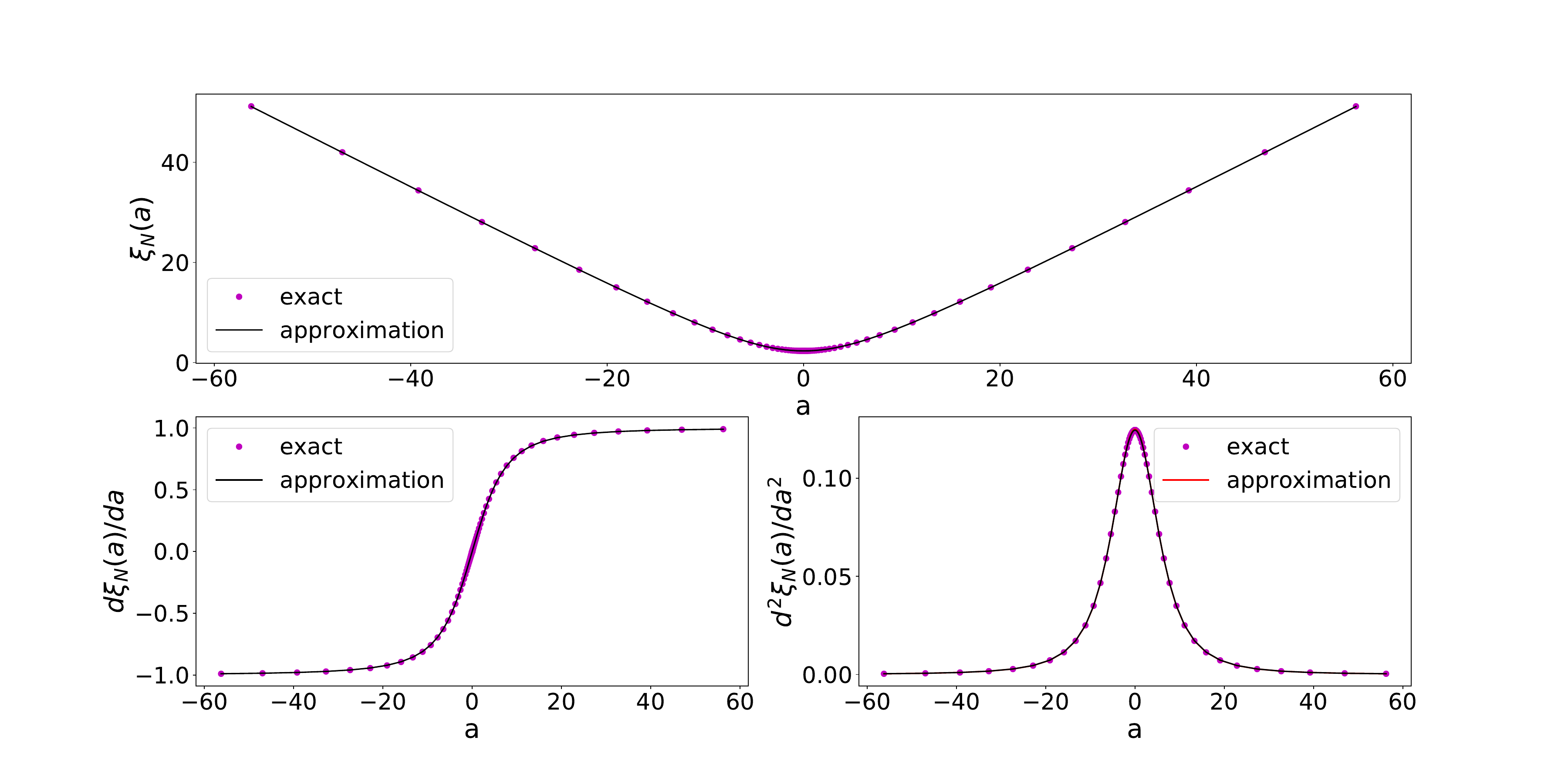}
\caption{ 
	 The approximation quality of  the formula \eqref{eq:normalh1formulax} for the function 
	$\xi_N(a)$ defined in \eqref{eq:defxi} and its first and second order differential.	
	{\bf Top: } the function $\xi_N$. {\bf Bottom left:} the first differential. 
	{\bf Bottom right:} the second differential. 
	Here $N=64$ but similar results are found for other values of $N$.
In all cases the exact value is computed using \eqref{eq:xi_as_dn}, \eqref{eq:dxida}
and a similar formula for the second order differential.	
}\label{fig:approximationqualityN64}
\end{figure}

 In order to derive relation ~\eqref{eq:normalh1formula} it is enough to prove
 ~\eqref{eq:normalh1formulax}. Note first that
 $E | x -Z |$ is asymptotically equal to $|x|$ for large values of $x$, which is also the case
 of the formula~\eqref{eq:normalh1formulax} (up to a constant which will not appear in the gradient). 
 Proving the approximation
amounts to analyzing the behavior of the function
 $\xi_N(a)$ for $a\simeq0$.
 We 
 consider 
 the second  derivative with respect to $a$ (the first derivative is zero because the leading term
is quadratic in $a$)
\beq \label{eq:ddf0}
\xi_N''(0) = \frac{1}{\sqrt{2}}
\frac{ \Gamma(\frac{N+1}{2})}{ \Gamma(1+\frac{N}{2})}. 
\eeq
It is now enough to see that we also have 
\beq
\xi_N(0)=(\sqrt{2}-1) \frac{\Gamma(\frac{N+1}{2})}{\Gamma(\frac{N}{2})},
\eeq 
to conclude that defining
\beq \label{eq:defcn01}
c_{N0}=\xi_N(0)- \frac{1}{\xi_N''(0)},  c_{N1}=\frac{1}{\xi_N''(0)^2},
\eeq
we have an approximation of $\xi_N(a)$ exact to second order in $a$ thus error of order $ O \left(
 | x|^4 \right)$ because of parity (the regularity is guaranteed by formula \eqref{eq:dxida}).
\begin{remark} \label{rem:sqrtxxplusN}
If one uses the  Stirling's formula in \eqref{eq:ddf0}
$\Gamma(z) = \sqrt{\frac{2 \pi}{z}} \left(\frac{z}{e}\right)^z (1+ O(1/z))$ it is possible to conclude, after straightforward computations,
that $\xi_N''(0) = \frac{1}{\sqrt{N}} +  O(1/N)$. This means that for some $c_1$ depending only on $N$:
\beqn \label{eq:normalh1formulaxcoarse}
& \ & d_{XH}(\delta_x,\normal{0}{I_N} )^2 = \sqrt{| x |^2 + N} - c_1
+ O \left(
 | x|^4 + \frac{1}{N} \right),
\eeqn
which is a coarser approximation and in practice gives less good results.
\end{remark} An illustration of the quality of the approximation \eqref{eq:normalh1formulax}
is given in figure \ref{fig:approximationqualityN64}.
\begin{remark}
Note that in \eqref{eq:ddf0}-\eqref{eq:defcn01} and similar approximations, some constants were explicitly written for numerical efficiency even if one can work up to a multiplicative constant. 
To take an example, 
consider the  factor $\Gamma\left(\frac{N+1}{2}\right)$ which is of order $(N/2)!$ (a large number in general). Removing it would make the numerical computation unstable; on the other hand we can exploit the reduction provided by the denominator which is here
$\Gamma\left(\frac{N}{2}\right)$; keeping both makes the overall coefficient $\frac{\Gamma\left(\frac{N+1}{2}\right)}{\Gamma\left(\frac{N}{2}\right)}$ numerically stable.
\end{remark}	
\section{Network architecture} \label{sec:architecture}
	
We follow in this section the specifications in ~\cite{CWAE} and reproduce below the corresponding architectures as
given in the reference:
	
\noindent {\bf MNIST/Fashion-MNIST} ($28\times28$ sized images): 

\textbf{encoder} three feed-forward ReLU layers, 200 neurons each;

\textbf{latent} $8$-dimensional;

\textbf{decoder} three feed-forward ReLU layers, 200 neurons each.

\noindent {\bf CIFAR-10} ($32\times 32$ images with $3$ color layers):

\textbf{encoder:} 
four convolution layers with $2\times2$ filters, the second one with $2\times2$ strides, other non-strided (3, 32, 32, and 32 channels) with ReLU activation, 128 ReLU neurons dense layer;

\textbf{latent} $64$-dimensional;

\textbf{decoder:}
 two dense ReLU layers with $128$ and $8192$ neurons,
two transposed-convolution layers with $2\times2$ filters (32 and 32 channels) and ReLU activation,
 a transposed convolution layer with $3\times3$ filter and $2\times2$ strides (32 channels) and ReLU activation,
 a transposed convolution layer with $2\times2$ filter (3 channels) and sigmoid activation.

The last layer returns the generated or reconstructed image.

All hyper-parameters are chosen as in the references: the loss was minimized with the Adam optimizer~\cite{kingma2014adam} with a learning rate of $0.001$  and hyper-parameters $\beta_1=0.9$, $\beta_2=0.999$, $\epsilon=10^{-8}$; we used $500$ epochs. 
The scaling parameter $\lambda$ was set to $100$.

\section{Implementation}
See attached notebook; 
the code is also available 
 as a Github repository at 
 
 \verb*|https://github.com/gabriel-turinici/radon_sobolev_vae|.

\subsection{Further numerical results} 
	We present here the result of the run corresponding to section~\ref{sec:num2terms} on the MNIST dataset when only
	the latent cost is present.
	
	\begin{figure}
		\includegraphics[width=.45\linewidth]{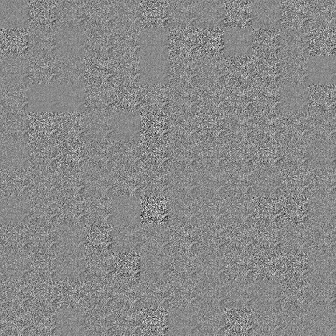}
		\includegraphics[width=.45\linewidth]{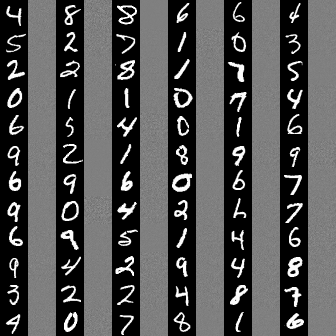}
		\caption{Results for section~\ref{sec:num2terms} on the MNIST dataset using only the latent term in the loss functional. The result is just white noise.
			{\bf  Left:}  generated samples.
			{\bf  Right:} reconstruction quality (latent only loss functional).
		}
		\label{fig:consistency}
	\end{figure}

\section{Additional details on the proofs and formulas}
For additional remarks on the proofs of Propositions~\ref{prop:onedim1h1} and~\ref{prop:formulash1},
two more approximations for $\gxn$ when $\Hcal = \dot{H}$ and 
 the computations of functions $\gxy$ and 
$\gxn$ for other choices of Sobolev spaces beyond $\dot{H}$ we refer the reader to \cite{turinici2020xray}.

\begin{thebibliography}{10}
	\expandafter\ifx\csname url\endcsname\relax
	\def\url#1{\texttt{#1}}\fi
	\expandafter\ifx\csname urlprefix\endcsname\relax\def\urlprefix{URL }\fi
	\expandafter\ifx\csname href\endcsname\relax
	\def\href#1#2{#2} \def\path#1{#1}\fi
	
	\bibitem{kingma2013autoencoding}
	D.~P. Kingma, M.~Welling, {Auto-Encoding Variational Bayes}, arxiv:1312.6114
	(2013).
	
	\bibitem{originalGAN14}
	I.~Goodfellow, J.~Pouget-Abadie, M.~Mirza, B.~Xu, D.~Warde-Farley, S.~Ozair,
	A.~Courville, Y.~Bengio,
	\href{http://papers.nips.cc/paper/5423-generative-adversarial-nets.pdf}{Generative
		adversarial nets}, in: Z.~Ghahramani, M.~Welling, C.~Cortes, N.~D. Lawrence,
	K.~Q. Weinberger (Eds.), Advances in Neural Information Processing Systems
	27, Curran Associates, Inc., 2014, pp. 2672--2680.
	\newline\urlprefix\url{http://papers.nips.cc/paper/5423-generative-adversarial-nets.pdf}
	
	\bibitem{tolstikhin2017wasserstein}
	I.~Tolstikhin, O.~Bousquet, S.~Gelly, B.~Schoelkopf, Wasserstein auto-encoders,
	arxiv:1711.01558 (2017).
	
	\bibitem{arjovsky2017wgan}
	M.~Arjovsky, S.~Chintala, L.~Bottou,
	\href{http://proceedings.mlr.press/v70/arjovsky17a.html}{{W}asserstein
		generative adversarial networks}, in: D.~Precup, Y.~W. Teh (Eds.),
	Proceedings of the 34th International Conference on Machine Learning, Vol.~70
	of Proceedings of Machine Learning Research, PMLR, International Convention
	Centre, Sydney, Australia, 2017, pp. 214--223.
	\newline\urlprefix\url{http://proceedings.mlr.press/v70/arjovsky17a.html}
	
	\bibitem{kingma_introduction_2019}
	D.~P. Kingma, W.~Max, An {Introduction} to {Variational} {Autoencoders}, Now
	Publishers Inc, 2019.
	
	\bibitem{radon_uber_1917}
	J.~Radon, Über die {Bestimmung} von {Funktionen} durch ihre {Integralwerte}
	längs gewisser {Mannigfaltigkeiten}, 1917, published: Leipz. Ber. 69,
	262-277 (1917).
	
	\bibitem{xraytransform38}
	F.~John, \href{https://doi.org/10.1215/S0012-7094-38-00423-5}{The
		ultrahyperbolic differential equation with four independent variables}, Duke
	Math. J. 4~(2) (1938) 300--322.
	\newblock \href {https://doi.org/10.1215/S0012-7094-38-00423-5}
	{\path{doi:10.1215/S0012-7094-38-00423-5}}.
	\newline\urlprefix\url{https://doi.org/10.1215/S0012-7094-38-00423-5}
	
	\bibitem{natterer_mathematics_1986}
	F.~Natterer, The mathematics of computerized tomography, 1986, published:
	Stuttgart: B. G. Teubner; Chichester etc.: John Wiley \& Sons. X, 222 p. DM
	72.00 (1986).
	
	\bibitem{book_Filippo}
	F.~{Santambrogio}, {Optimal transport for applied mathematicians. Calculus of
		variations, PDEs, and modeling.}, Vol.~87, Cham: Birkh\"auser/Springer, 2015.
	
	\bibitem{giglibook}
	L.~{Ambrosio}, N.~{Gigli}, G.~{Savar\'e}, {Gradient flows in metric spaces and
		in the space of probability measures. 2nd ed.}, 2nd Edition, Basel:
	Birkh\"auser, 2008.
	
	\bibitem{jordan_neuman_inner_product}
	P.~Jordan, J.~V. Neumann, \href{http://www.jstor.org/stable/1968653}{On inner
		products in linear, metric spaces}, Annals of Mathematics 36~(3) (1935)
	719--723.
	\newline\urlprefix\url{http://www.jstor.org/stable/1968653}
	
	\bibitem{blumenthal_theory_1953}
	L.~M. Blumenthal, Theory and applications of distance geometry, 1953,
	published: Oxford: At the Clarendon Press (Geoffrey Cumberlege), XI, 347 p.
	(1953).
	
	\bibitem{day_criteria_1959}
	M.~M. Day, \href{http://www.jstor.org/stable/2032894}{On {Criteria} of
		{Kasahara} and {Blumenthal} for {Inner}-{Product} {Spaces}}, Proceedings of
	the American Mathematical Society 10~(1) (1959) 92--100, publisher: American
	Mathematical Society.
	\newblock \href {https://doi.org/10.2307/2032894} {\path{doi:10.2307/2032894}}.
	\newline\urlprefix\url{http://www.jstor.org/stable/2032894}
	
	\bibitem{Kolouri_2018_CVPR}
	S.~Kolouri, G.~K. Rohde, H.~Hoffmann, {Sliced Wasserstein Distance for Learning
		Gaussian Mixture Models}, in: The IEEE Conference on Computer Vision and
	Pattern Recognition (CVPR), 2018.
	
	\bibitem{Deshpande_2018_CVPR}
	I.~Deshpande, Z.~Zhang, A.~G. Schwing, {Generative Modeling Using the Sliced
		Wasserstein Distance}, in: The IEEE Conference on Computer Vision and Pattern
	Recognition (CVPR), 2018.
	
	\bibitem{kolouri2018sliced}
	S.~Kolouri, P.~E. Pope, C.~E. Martin, G.~K. Rohde,
	\href{https://openreview.net/forum?id=H1xaJn05FQ}{{Sliced Wasserstein
			Auto-Encoders}}, in: International Conference on Learning Representations,
	2019.
	\newline\urlprefix\url{https://openreview.net/forum?id=H1xaJn05FQ}
	
	\bibitem{Wu_2019_CVPRgenerative}
	J.~Wu, Z.~Huang, D.~Acharya, W.~Li, J.~Thoma, D.~P. Paudel, L.~V. Gool, Sliced
	{W}asserstein generative models, in: The IEEE Conference on Computer Vision
	and Pattern Recognition (CVPR), 2019.
	
	\bibitem{Deshpande_2019_CVPRmax}
	I.~Deshpande, Y.-T. Hu, R.~Sun, A.~Pyrros, N.~Siddiqui, S.~Koyejo, Z.~Zhao,
	D.~Forsyth, A.~G. Schwing, {Max-Sliced Wasserstein Distance and Its Use for
		GANs}, in: The IEEE Conference on Computer Vision and Pattern Recognition
	(CVPR), 2019.
	
	\bibitem{CWAE}
	J.~Tabor, S.~Knop, P.~Spurek, I.~T. Podolak, M.~Mazur, S.~Jastrz{k{e}}bski,
	\href{http://arxiv.org/abs/1805.09235}{{Cramer-Wold AutoEncoder}}, CoRR
	abs/1805.09235 (2018).
	\newblock \href {http://arxiv.org/abs/1805.09235} {\path{arXiv:1805.09235}}.
	\newline\urlprefix\url{http://arxiv.org/abs/1805.09235}
	
	\bibitem{SZEKELY13}
	G.~J. Szekely, M.~L. Rizzo,
	\href{http://www.sciencedirect.com/science/article/pii/S0378375813000633}{Energy
		statistics: {A} class of statistics based on distances}, Journal of
	Statistical Planning and Inference 143~(8) (2013) 1249--1272.
	\newblock \href {https://doi.org/10.1016/j.jspi.2013.03.018}
	{\path{doi:10.1016/j.jspi.2013.03.018}}.
	\newline\urlprefix\url{http://www.sciencedirect.com/science/article/pii/S0378375813000633}
	
	\bibitem{sobolevgan}
	Y.~Mroueh, C.-L. Li, T.~Sercu, A.~Raj, Y.~Cheng,
	\href{https://openreview.net/forum?id=SJA7xfb0b}{{Sobolev GAN}}, in:
	International Conference on Learning Representations, 2018.
	\newline\urlprefix\url{https://openreview.net/forum?id=SJA7xfb0b}
	
	\bibitem{cramergan}
	M.~G. Bellemare, I.~Danihelka, W.~Dabney, S.~Mohamed, B.~Lakshminarayanan,
	S.~Hoyer, R.~Munos, \href{http://arxiv.org/abs/1705.10743}{{The Cramer
			Distance as a Solution to Biased Wasserstein Gradients}}, CoRR abs/1705.10743
	(2017).
	\newblock \href {http://arxiv.org/abs/1705.10743} {\path{arXiv:1705.10743}}.
	\newline\urlprefix\url{http://arxiv.org/abs/1705.10743}
	
	\bibitem{hilbertspaceembed10}
	B.~K. {Sriperumbudur}, A.~{Gretton}, K.~{Fukumizu}, B.~{Sch\"olkopf}, G.~R.~G.
	{Lanckriet}, {Hilbert space embeddings and metrics on probability measures},
	{J. Mach. Learn. Res.} 11 (2010) 1517--1561.
	
	\bibitem{AnnalsStat_endist13}
	D.~Sejdinovic, B.~Sriperumbudur, A.~Gretton, K.~Fukumizu,
	\href{http://www.jstor.org/stable/23566550}{Equivalence of distance-based and
		{RKHS}-based statistics in hypothesis testing}, The Annals of Statistics
	41~(5) (2013) 2263--2291.
	\newline\urlprefix\url{http://www.jstor.org/stable/23566550}
	
	\bibitem{karras2018progressive}
	T.~Karras, T.~Aila, S.~Laine, J.~Lehtinen,
	\href{https://openreview.net/forum?id=Hk99zCeAb}{Progressive growing of
		{GAN}s for improved quality, stability, and variation}, in: International
	Conference on Learning Representations, 2018.
	\newline\urlprefix\url{https://openreview.net/forum?id=Hk99zCeAb}
	
	\bibitem{adamssobolev}
	R.~A. {Adams}, J.~J.~F. {Fournier}, {Sobolev spaces. 2nd ed.}, 2nd Edition, New
	York, NY: Academic Press, 2003.
	
	\bibitem{lionshmoins1}
	J.~Deny, J.-L. Lions, \href{http://www.numdam.org/item/AIF_1954__5__305_0}{Les
		espaces du type de {Beppo Levi}}, Annales de l'Institut Fourier 5 (1954)
	305--370.
	\newblock \href {https://doi.org/10.5802/aif.55} {\path{doi:10.5802/aif.55}}.
	\newline\urlprefix\url{http://www.numdam.org/item/AIF_1954__5__305_0}
	
	\bibitem{rpeyre}
	{Peyre, R\'emi}, \href{https://doi.org/10.1051/cocv/2017050}{{Comparison
			between W2 distance and H1 norm, and Localization of Wasserstein distance}},
	ESAIM: COCV 24~(4) (2018) 1489--1501.
	\newblock \href {https://doi.org/10.1051/cocv/2017050}
	{\path{doi:10.1051/cocv/2017050}}.
	\newline\urlprefix\url{https://doi.org/10.1051/cocv/2017050}
	
	\bibitem{ottovillani2000}
	F.~{Otto}, C.~{Villani}, {Generalization of an inequality by Talagrand and
		links with the logarithmic Sobolev inequality.}, {J. Funct. Anal.} 173~(2)
	(2000) 361--400.
	
	\bibitem{brasco2021characterisation}
	L.~Brasco, D.~Gómez-Castro, J.~L. Vázquez, Characterisation of homogeneous
	fractional sobolev spaces (2021).
	\newblock \href {http://arxiv.org/abs/2007.08000} {\path{arXiv:2007.08000}}.
	
	\bibitem{fid_metric}
	M.~Heusel, H.~Ramsauer, T.~Unterthiner, B.~Nessler, S.~Hochreiter,
	\href{https://proceedings.neurips.cc/paper/2017/file/8a1d694707eb0fefe65871369074926d-Paper.pdf}{Gans
		trained by a two time-scale update rule converge to a local nash
		equilibrium}, in: I.~Guyon, U.~V. Luxburg, S.~Bengio, H.~Wallach, R.~Fergus,
	S.~Vishwanathan, R.~Garnett (Eds.), Advances in Neural Information Processing
	Systems, Vol.~30, Curran Associates, Inc., 2017, pp. 6626--6637.
	\newline\urlprefix\url{https://proceedings.neurips.cc/paper/2017/file/8a1d694707eb0fefe65871369074926d-Paper.pdf}
	
	\bibitem{noncentralchi}
	D.~{Kerridge}, {A probabilistic derivation of the non-central \(\chi^ 2\)
		distribution}, {Aust. J. Stat.} 7 (1965) 37--39.
	
	\bibitem{johnson_kotz_stat_book}
	N.~L. {Johnson}, S.~{Kotz}, N.~{Balakrishnan}, {Continuous univariate
		distributions. Vol. 2. 2nd ed}, 2nd Edition, New York, NY: Wiley, 1995.
	
	\bibitem{kingma2014adam}
	D.~P. Kingma, J.~Ba, Adam: A method for stochastic optimization,
	arxiv:1412.6980 (2014).
	
	\bibitem{turinici2020xray}
	G.~Turinici, X-ray {S}obolev variational auto-encoders (2020).
	\newblock \href {http://arxiv.org/abs/1911.13135} {\path{arXiv:1911.13135}}.
	
\end{thebibliography}
\end{document}